\documentclass[10pt,twocolumn,letterpaper]{article}

\usepackage[pagenumbers]{cvpr} 

\usepackage{graphicx}
\usepackage{amsmath}
\usepackage{amssymb}
\usepackage{booktabs}

\usepackage{multirow}

\usepackage[pagebackref,breaklinks,colorlinks]{hyperref}

\usepackage{breakurl}

\usepackage[capitalize]{cleveref}
\crefname{section}{Sec.}{Secs.}
\Crefname{section}{Section}{Sections}
\Crefname{table}{Table}{Tables}
\crefname{table}{Tab.}{Tabs.}

\def\cvprPaperID{*****} 
\def\confName{CVPR}
\def\confYear{2023}

\begin{document}

\title{Trained Latent Space Navigation to Prevent Lack of Photorealism in Generated Images on Style-based Models}

\author{Takumi Harada${}^{1}$, Kazuyuki Aihara${}^{2}$, Hiroyuki Sakai${}^{1}$\\
${}^{1}$TOYOTA CENTRAL R\&D LABS., INC.   ${}^{2}$University of Tokyo\\
{\tt\small t-harada@mosk.tytlabs.co.jp, kaihara@g.ecc.u-tokyo.ac.jp, sakai@mosk.tytlabs.co.jp}}

\maketitle

\begin{abstract}
Recent studies on StyleGAN variants show promising performances for various generation tasks. In these models, latent codes have traditionally been manipulated and searched for the desired images. However, this approach sometimes suffers from a lack of photorealism in generated images due to a lack of knowledge about the geometry of the trained latent space. In this paper, we show a simple unsupervised method that provides well-trained local latent subspace, enabling latent code navigation while preserving the photorealism of the generated images. Specifically, the method identifies densely mapped latent spaces and restricts latent manipulations within the local latent subspace. Experimental results demonstrate that images generated within the local latent subspace maintain photorealism even when the latent codes are significantly and repeatedly manipulated. Moreover, experiments show that the method can be applied to latent code optimization for various types of style-based models. Our empirical evidence of the method will benefit applications in style-based models.

\end{abstract}

\section{Introduction}
\label{sec:intro}
Generative adversarial networks (GANs) \cite{NIPS2014_5ca3e9b1} have shown impressive results in generating photo-realistic images. In particular, the StyleGAN architecture \cite{Karras_2019_CVPR, Karras_2020_CVPR, NEURIPS2021_076ccd93} has achieved fascinating results when editing high-quality images. An important property of StyleGAN is its ability to train latent code distribution. The trained latent code distribution shows a better match with the distribution of the training data. It has a disentanglement property: the semantic separability of generated images in latent space. Because of this attractive property, many strategies for latent code manipulation \cite{Shen_2020_CVPR, NEURIPS2020_6fe43269, Shen_2021_CVPR, Jahanian2020On, Li_2021_CVPR}, other synthesis tasks \cite{Skorokhodov_2022_CVPR, Shi_2022_CVPR, Chan_2022_CVPR}, and human interaction applications \cite{tejeda2020improving, liao2021interactive} have been studied.

Despite these successes, it is still challenging to manipulate latent codes significantly while maintaining photorealism. For example, latent search methods to obtain the user's desired image have been studied in human interaction applications \cite{tejeda2020improving, liao2021interactive}. In such applications, users require a large traversal in latent space since users often search for images of different identities and may change their desired image during the searching process. However, the large traversal in latent space sometimes leaves the trained latent space and generates images that lack photorealism. This out-of-distribution problem arises from a lack of knowledge about the geometry of the trained latent space when updating the latent code.

\begin{figure}[t]
  \centering
   \includegraphics[width=1.0\linewidth]{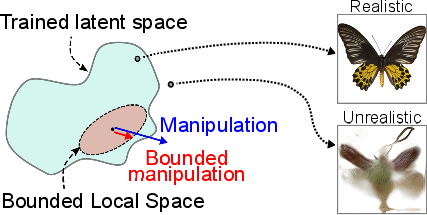}
    \caption{{\bf Overview.} Our method restricts updated latent codes to remain inside Bounded Local Space, a latent subspace that is densely mapped by the mapping network of the style-based model. Consequently, our method allows latent code manipulation within the trained latent space while preserving the reality of the generated images.}
   \label{fig:concept}
\end{figure}

Traditionally, to avoid exiting the trained latent space, image editing that changes attributes of a source image is implemented by manually setting a small step size for latent code manipulation \cite{Shen_2020_CVPR, Li_2021_CVPR, choi2022do}, applying a regularization term for latent codes \cite{Patashnik_2021_ICCV}, or utilizing an additional network that estimates differences from the source image \cite{Patashnik_2021_ICCV, Wu_2021_CVPR, Saha_2021_CVPR}. These measures are suitable for image attribute editing that requires small-traversal manipulation of the latent codes but not for large traversals. Small-traversal manipulation requires more iterations to achieve large traversals, making it difficult for users to use. 

Another line of research has tried to correct latent codes to remain inside the trained latent space. For example, Shen \emph{et al.} \cite{Shen_2020_CVPR} trained a linear support-vector machine using 4,000 manually labeled bad images and manipulated the latent codes of the collapsed images to the direction of good images using the trained hyperplane. In another approach, Wen \emph{et al.} \cite{https://doi.org/10.48550/arxiv.2104.05518} moved latent codes of collapsed images onto the hypersphere of the latent space via three optimization steps. These methods, however, are time-consuming because it is necessary to prepare an additional training dataset, train an additional classifier, and compute multistage optimization. In addition, these methods primarily focus on recovery from collapsed images and require human visual inspection to determine if the methods are necessary. As a result of these shortcomings, these methods are unsuitable for combination with latent code manipulation, especially in loops that are performed iteratively and automatically. Thus, knowledge about the geometry of trained latent space, which preserves the reality of generated images even if the latent code is moved significantly in a single update or moved repeatedly, is helpful for applications that require large traversals and allows efficient broad latent searches. However, a verified method and empirical evidence for identifying trained latent space are still lacking.

A simple and easy way to obtain partial knowledge about trained latent space is to identify the local manifold extracted from a mapping network of a style-based model in an unsupervised manner. A recent study reports that small steps directed to one of the singular vectors obtained from the Jacobian matrix of a mapping network have robustness for the photorealism of the generated images \cite{choi2022do}. Additionally, the study reports the robustness in the subspace spanned by two singular vectors. While these findings are useful for the photorealism of the generated images, they provide only limited spatial (two dimensions or less) evidence for a large latent space. Furthermore, the extent to which a latent code can be moved in a single update while preserving photorealism remains unknown. Given the variety of applications, knowledge of a broader (higher-dimensional) latent space is preferable for generating various images. In addition, considering large traversal applications, clarifying how far a latent code can be moved in a single update is essential for reducing the number of computational iterations.

In this study, we investigate generated image photorealism in higher-dimensional subspaces of the latent space and the relationship between subspace size and generated image photorealism. As a complement to the findings of the recent study \cite{choi2022do}, we provide empirical evidence that a latent space traversal within Bounded Local Space, which is spanned by singular vectors and limited in extent by the singular values in each direction of the singular vectors, for style-based models (called here ``our method'' for simplicity) can preserve the photorealism of the generated images even in the large traversal of latent space. Our critical insight is that singular values and vectors obtained from the Jacobian matrix of a mapping network in style-based models can navigate trained latent space exploration. Unlike previous studies on latent code correction \cite{Shen_2020_CVPR, https://doi.org/10.48550/arxiv.2104.05518}, our method is not computationally expensive, does not require any additional datasets or training, and can dynamically traverse a trained latent space.

The main contributions of this study can be summarized as follows:
\begin{itemize}
\item Empirical evidence that a latent space traversal within Bounded Local Space preserves photorealism even if the latent code is moved significantly in a single update and moved repeatedly;
\item Experiments showing that a latent space traversal within Bounded Local Space can be adapted to style-based models that have a mapping network, even if the models are trained on different datasets; and
\item Experiments showing that a latent space traversal within Bounded Local Space can be applied to latent code manipulation through optimization with different types of loss functions.
\end{itemize}

\section{Related Work}
\label{sec:related}

\subsection{Style-based Models}
Generative adversarial networks have shown a good ability to generate photo-realistic images \cite{DBLP:journals/corr/RadfordMC15, karras2018progressive, brock2018large}. Specifically, StyleGAN \cite{Karras_2019_CVPR}, a style-based model, opens the door to high-quality image generation and editing. A style-based model consists of a mapping network and a synthesis network. A mapping network converts a Gaussian-distributed latent code $\mathbf{z} \in \mathcal{Z}$ into an intermediate latent code $\mathbf{w} \in \mathcal{W}$ that better matches the training data distribution. The intermediate latent code $\mathbf{w}$, which controls the style of the output image, is inserted into the synthesis network, and the synthesis network subsequently generates an image. The intermediate latent space $\mathcal{W}$ provides considerable disentanglement properties beneficial for semantic image editing \cite{Karras_2019_CVPR, 9241434}. This attractive feature has inspired many application studies: for example, video synthesis \cite{Skorokhodov_2022_CVPR}, compositional image synthesis \cite{Shi_2022_CVPR}, and novel view synthesis \cite{Chan_2022_CVPR}. Even though it is an active research area, little work has been done concerning the latent space large traversal that preserves the photorealism of the generated images.

\subsection{Recovery from Collapsed Images}
Some studies have proposed latent correction techniques to restore a collapsed image to a realistic image. Shen \emph{et al.} \cite{Shen_2020_CVPR} manipulate the latent codes of collapsed images to be in the direction of good images using a support-vector machine. While image artifacts are corrected, this method requires additional data preparation for bad-quality images and training. Inherently, collecting bad-quality images over a large area of the latent space is difficult. Wen \emph{et al.} \cite{https://doi.org/10.48550/arxiv.2104.05518} move the latent codes of collapsed images onto the hypersphere of the latent space via three optimization steps. While achieving recovery from collapsed images, this method requires a time-consuming multistage optimization.

In essence, previous methods for recovering collapsed images are unnecessary if the manipulated latent codes do not deviate from the trained latent space. In this regard, our method prevents deviations from the trained latent space by limiting the range of the latent code manipulation. In addition, our method has the following features that previous methods do not: no additional data preparation or training is required, and there is a low computational cost. Therefore, our method allows for large traversals without requiring collapsed image recovery.

\subsection{Latent Code Manipulation}
In latent code manipulation, attribute editing requiring small traversals has been the leading research topic, while large traversals have been overlooked. This section outlines latent code manipulations requiring small traversals and shows the relationship between our method and the previous method we specifically refer to.

Recent studies have primarily focused on identifying the desired edits' manipulation direction in the latent space. Several studies have investigated supervised methods that use extra-supervised learning models to guide the manipulation direction in the latent space \cite{Shen_2020_CVPR, 9010379, Patashnik_2021_ICCV, 1011453447648, Li_2021_CVPR, Jahanian2020On}. Another line of study has generated unsupervised methods that explore the manipulation direction from the information inside a trained model \cite{choi2022do, https://doi.org/10.48550/arxiv.1812.01161, NEURIPS2021_8b406655, Shen_2021_CVPR, NEURIPS2020_6fe43269}. For example, SeFa \cite{Shen_2021_CVPR} uses eigenvectors obtained from the first affine layer for $\mathcal{W}$; GANSpace \cite{NEURIPS2020_6fe43269} uses eigenvectors obtained from the activation space of the network using principal component analysis; and Local Basis \cite{choi2022do} uses singular vectors obtained from the Jacobian matrix of a mapping network.

Even though previous methods are useful for image editing, they do not examine how the trained latent space is spread out and distributed. In this regard, the Iterative-Curve Traversal (ICT) method \cite{choi2022do}, an iterative application of the Local Basis manipulation with small steps, demonstrates robustness for image quality due to tracing to the latent manifold. The robustness is examined on an axis directed to one of the singular vectors and in the subspace spanned by two singular vectors. However, the robustness in the higher-dimensional subspace spanned by singular vectors and the robustness in terms of the manipulation magnitude in a single update remains uninvestigated. We newly define a high-dimensional local subspace using singular values and vectors (called Bounded Local Space), which is highly motivated by ICT, and add empirical evidence for the robustness in Bounded Local Space with the manipulation magnitude. Based on the evidence, Bounded Local Space can be used for various applications. Specifically, Bounded Local Space dynamically restricts the manipulation's magnitude, includes a higher-dimensional subspace, and is, therefore, better suited for large traversals in various directions. Note that our aim in this study is to demonstrate the photorealism preservation in Bounded Local Space for the large traversals and not to claim that latent code manipulation within Bounded Local Space is a better attribute editing method because Bounded Local Space is a method to constrain the amount of latent code movement in a single update.

\section{Method}
\label{sec:method}
Our goal is to achieve a latent code manipulation that does not deviate from the trained latent space. Accordingly, we propose Bounded Local Space that enables the trained local latent space to be navigated (Fig. \ref{fig:concept}). The main idea is to extract information about the local latent space that the mapping network densely projects. In the following, we discuss Bounded Local Space (\ref{subsec:BLS}) and a latent code manipulation algorithm within Bounded Local Space (\ref{subsec:BLS2}).

\subsection{Bounded Local Space}
\label{subsec:BLS}
A typical GAN generator maps a latent vector $\mathbf{z}$ in the input latent space $\mathcal{Z} \subseteq \mathbb{R}^{n}$ to an image $\mathbf{x}$ in the image space $\mathcal{X} \subseteq \mathbb{R}^{W\times H\times C}$. In a style-based model, a latent code $\mathbf{z}$ is first mapped to an intermediate latent code $\mathbf{w} \in \mathcal{W} \subseteq \mathbb{R}^{n}$ by a mapping network $M: \mathcal{Z} \rightarrow \mathcal{W}$. A synthesis network in a style-based model $G: \mathcal{W} \rightarrow \mathcal{X}$ then maps the intermediate latent code $\mathbf{w}$ to an image $\mathbf{x}$. In this generation process, Bounded Local Space functions as a gatekeeper to prevent movement into an untrained latent space in the $\mathcal{W}$ space.

Bounded Local Space is defined as a bounded subspace spanned by the Local Basis \cite{choi2022do}. Accordingly, we first define the Local Basis. We define a tangent space at $\mathbf{w}$, denoted by $\mathcal{T_{\mathbf{w}}W}$. The tangent space is a set of tangent vectors through point $\mathbf{w}$. To obtain the basis of the tangent space, called the Local Basis \cite{choi2022do}, we utilize the differential of a mapping network $dM_{\mathbf{z}}: \mathcal{T_{\mathbf{z}}Z} \rightarrow \mathcal{T_{\mathbf{w}}W}$ that linearly maps the tangent space at $\mathbf{z}$, denoted by $\mathcal{T_{\mathbf{z}}Z}$, to the tangent space at $\mathbf{w}$, where $\mathbf{w} = M(\mathbf{z})$. The Local Basis is obtained as the left singular vectors via singular value decomposition of the Jacobian matrix of a mapping network $\mathbf{J}$ that is the matrix representation of $dM_{\mathbf{z}}$.

\begin{equation}
\mathbf{J} = \frac{\partial M(\mathbf{z})}{\partial \mathbf{z}} = \mathbf{U\Sigma V}^{T}
\label{eq:1}
\end{equation}
\begin{equation}
dM_{\mathbf{z}}(\mathbf{v}_{i}^{\mathbf{z}}) = \sigma_{i}^{\mathbf{z}}\mathbf{u}_{i}^{\mathbf{w}}
\label{eq:2}
\end{equation}
\begin{equation}
Local Basis(\mathbf{w} = M(\mathbf{z})) = \left \{ \mathbf{u}_{i}^{\mathbf{w}} \right \}_{i = 1}^{n}
\label{eq:3}
\end{equation}
Here, $\mathbf{U} = [ \mathbf{u}_{1}^{\mathbf{w}}, \cdots , \mathbf{u}_{n}^{\mathbf{w}} ] \in \mathbb{R}^{n\times n}$ is the matrix of the left singular vectors, $\mathbf{u}_{i}^{\mathbf{w}}$ is the $i$-th left singular vector, $\mathbf{\Sigma} = diag( \sigma_{1}^{\mathbf{z}}, \cdots , \sigma_{n}^{\mathbf{z}} ) \in \mathbb{R}^{n \times n}$ is the diagonal matrix of the singular values, $\sigma_{i}^{\mathbf{z}}$ is the $i$-th singular value with $\sigma_{1}^{\mathbf{z}} \geq \cdots \geq \sigma_{n}^{\mathbf{z}}$, $\mathbf{V} = [ \mathbf{v}_{1}^{\mathbf{z}}, \cdots , \mathbf{v}_{n}^{\mathbf{z}} ] \in \mathbb{R}^{n \times n}$ is the matrix of the right singular vectors, and $\mathbf{v}_{i}^{\mathbf{z}}$ is the $i$-th right singular vector.

Therefore, Bounded Local Space is defined as follows.
\begin{equation}
\begin{aligned}
Boun&ded Local Space(\mathbf{w} = M(\mathbf{z})) = \\
&\{ \mathbf{w} + \sum_{i=1}^{n}\lambda_{i}\mathbf{u}_{i}^{\mathbf{w}} | \lambda_{i}\in [-\alpha\sigma_{i}^{\mathbf{z}},\alpha\sigma_{i}^{\mathbf{z}}], \alpha \in \mathbb{R}^{+}\}
\label{eq:4}
\end{aligned}
\end{equation}
Here, $\lambda_{i}$ is a bounding factor that uses the singular values of the Jacobian matrix to limit the magnitude in each direction of the Local Basis, and $\alpha$ is a scaling factor controls the movable range of the subspace.

\subsection{Manipulation within Bounded Local Space}
\label{subsec:BLS2}

\begin{figure}[tb]
  \centering

   \includegraphics[width=1.0\linewidth]{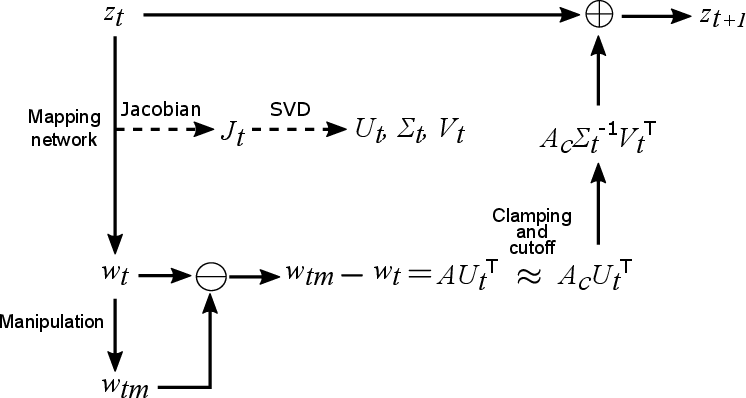}

    \caption{{\bf Latent Space Traversal Algorithm.} First, singular value decomposition (SVD) is applied to the Jacobian matrix for the latent code $\mathbf{z}_t$. Second, the target latent code $\mathbf{w}_{tm}$ is obtained. Third, the change in the latent code in $\mathcal{W}$ space is represented using a coefficient vector $\mathbf{A}$ and left singular vectors $\mathbf{U}_t$. Fourth, the coefficient vector is clamped and is partly cut off using the singular values $\mathbf{\Sigma}_t$. Fifth, the change in the latent code in $\mathcal{W}$ space is converted into the change in the latent code in $\mathcal{Z}$ space. Finally, the latent code $\mathbf{z}_t$ is updated to $\mathbf{z}_{t+1}$.}
   \label{fig:process}
\end{figure}

To manipulate latent codes within Bounded Local Space, we further introduce a latent space traversal algorithm (Fig. \ref{fig:process}). Briefly, the latent space traversal algorithm is an iterative method that computes Bounded Local Space at each step of the latent manipulation and restricts the movement of the updated latent codes to the interior of Bounded Local Space. Let $\mathbf{z}_{t}$ and $\mathbf{w}_{t}$ be current latent codes, where $\mathbf{w}_{t} = M(\mathbf{z}_{t})$. We assume we have a target latent code $\mathbf{w}_{tm}$ that is obtained from, for example, an image editing method or optimization using an external network. Then, we calculate the Local Basis at $\mathbf{w}_{t}$ and obtain the coefficient vector $\mathbf{A} \in \mathbb{R}^{1\times n}$ for the Local Basis:
\begin{equation}
\Delta \mathbf{w} = \mathbf{w}_{tm} - \mathbf{w}_{t} = \mathbf{AU}_{t}^{T},
\label{eq:6}
\end{equation}
where $\mathbf{A} = [ a_{1}, \cdots ,a_{n} ]$ is the coefficient vector whose element determines the magnitude of each Local Basis, and $\mathbf{U}_t = [ \mathbf{u}_{1}^{\mathbf{w}_t}, \cdots , \mathbf{u}_{n}^{\mathbf{w}_t} ]$ is the matrix of the left singular vectors of the Jacobian matrix of the mapping network at $\mathbf{z}_{t}$. The coefficient vector $\mathbf{A}$ is obtained using the following equation: 
\begin{equation}
\mathbf{A} = \Delta \mathbf{w} (\mathbf{U}_{t}^{T})^{-1}.
\label{eq:7}
\end{equation}
To restrict the movement of the updated latent codes to the interior of Bounded Local Space, we clamp each element $a_{i}$ of $\mathbf{A}$ with the corresponding singular value $\sigma_{i}^{\mathbf{z}_t}$ and obtain a clamped coefficient vector $\mathbf{A}_{c} = [ a_{c1}, \cdots , a_{cn} ]$, where $a_{ci} \in [ -\alpha\sigma_{i}^{\mathbf{z}_t}, \alpha\sigma_{i}^{\mathbf{z}_t} ]$. 
\begin{equation}
\Delta \mathbf{w} \approx \mathbf{A_{c}U}_{t}^T
\label{eq:8}
\end{equation}
Then, we update the current latent code toward the target latent code in $\mathcal{Z}$ space, assuming $\mathbf{w}_t+\mathbf{u}_{i}^{\mathbf{w}_t} \approx M(\mathbf{z}_t+\sigma_{i}^{\mathbf{z}_{t}-1}\mathbf{v}_{i}^{\mathbf{z}_t})$.
\begin{equation}
\mathbf{z}_{t+1} = \mathbf{z}_{t} + \mathbf{A_{c}\Sigma}_{t}^{-1}\mathbf{V}_{t}^T
\label{eq:9}
\end{equation}
\begin{equation}
\mathbf{w}_{t+1} = M(\mathbf{z}_{t+1})
\label{eq:10}
\end{equation}
Here, $\mathbf{\Sigma}_t = diag( \sigma_{1}^{\mathbf{z}_t}, \cdots , \sigma_{n}^{\mathbf{z}_t} )$ is the diagonal matrix of the singular values of the Jacobian matrix of the mapping network at $\mathbf{z}_{t}$, and $\mathbf{V}_t = [ \mathbf{v}_{1}^{\mathbf{z}_t}, \cdots and \mathbf{v}_{n}^{\mathbf{z}_t} ]$ is the matrix of the right singular vectors of the Jacobian matrix of the mapping network at $\mathbf{z}_{t}$. By repeating the above process, this method can also be applied within the optimization loop that iteratively changes the target latent code. Note that, to calculate the Local Basis at $\mathbf{w}$, we need the corresponding latent code $\mathbf{z}$ and, therefore, to update the latent codes in the $\mathcal{Z}$ space.

\section{Experiments}
\label{sec:exp}
We conduct a latent traversal experiment to investigate the photorealism within Bounded Local Space (\ref{subsec:lat}). In this experiment, we evaluate the robustness of the photorealism when the latent code is moved repeatedly and the direction of latent code movement to confirm that it is moving in the target direction. In addition, to investigate how significant manipulation's magnitude in a single update can preserve the photorealism in our method, we evaluate the impact of the scaling factor on the robustness and the direction of latent code movement. Furthermore, we present the results of applying our method to optimization problems (\ref{subsec:app}) since application to optimization problems opens the door to various applications. Implementation details can be found in Supplementary Material Section 1.

\noindent {\bf Models and Datasets:} Here, we use four types of style-based models: StyleGAN2 \cite{Karras_2020_CVPR}, StyleGAN3 \cite{NEURIPS2021_076ccd93}, SemanticStyleGAN \cite{Shi_2022_CVPR}, and Efficient Geometry-aware 3D Generative Adversarial Networks (EG3D) \cite{Chan_2022_CVPR}. For StyleGAN2, we use two models trained on the LHQ dataset \cite{Skorokhodov_2021_ICCV} and the butterfly dataset \cite{butterfly}, respectively. For StyleGAN3, we use two models trained on the FFHQ dataset \cite{Karras_2019_CVPR} and the WikiArt dataset \cite{wikiart}, respectively. SemanticStyleGAN is trained on CelebAMask-HQ \cite{CelebAMask-HQ}, and EG3D is trained on AFHQv2 Cats \cite{9157662, NEURIPS2021_076ccd93}.

\noindent {\bf Metrics:} We use the Fr\'{e}chet Inception Distance (FID) score \cite{NIPS2017_8a1d6947} to quantify the image quality for photorealism. To evaluate if an updated latent code is moving toward a target direction, we use the cosine similarity between the direction of the updated latent code and the target direction and the cumulative distance traveled in the target direction.

\begin{figure}[tb]
  \centering
   \includegraphics[width=1.0\linewidth]{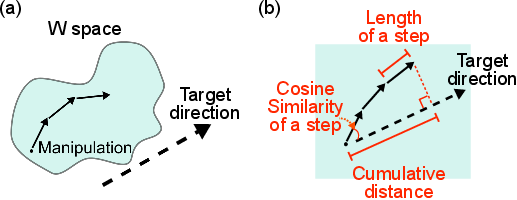}
\caption{{\bf Latent Traversal Experiment.} (a) Illustration of the latent traversal experiment. First, the initial latent code and the target direction are each set randomly. Then, the initial latent code is updated 500 times in the target direction. This process is performed for 1,000 randomly set pairs of initial latent codes and target directions. (b) Evaluation metrics for the traversal efficiency. Cosine similarity confirms that the updated latent code direction is oriented toward the target direction. The length of a step checks if our method dynamically adjusts the size of the updates. The cumulative distance checks if the updated latent code is advancing in the target direction.}
   \label{fig:fid}
\end{figure}

\subsection{Latent Traversal}
\label{subsec:lat}
To examine the photorealism of our method, we conduct a latent traversal experiment using various types of StyleGAN models. Specifically, we randomly set an initial latent code and a target direction and move the latent code in the target direction in $\mathcal{W}$ space (Fig. \ref{fig:fid}a). The latent code is moved for 500 iterations, with the distance traveled in each iteration being a specific Euclidean distance. The distance is much larger than that in previous latent code manipulations for attribute editing (see Supplementary Material Section 2). Note that the distance is bounded by Bounded Local Space in our method. This procedure is performed on 1,000 randomly determined pairs of initial latent codes and target directions. For comparison, in addition to our method, latent codes are updated via Linear traversal, Random traversal, and ICT \cite{choi2022do}. We evaluate Linear traversal, which simply moves the latent code in the target direction, to show that updating the latent code in one direction easily deviates from the trained latent space. We evaluate Random traversal, which moves the latent code in a random direction (a positive or negative sign is changed to face the target direction), to show that being in the trained latent space by moving with our method is not a coincidence. We evaluate ICT \cite{choi2022do}, which moves the latent code to one Local Basis $ \mathbf{u}_{i}^{\mathbf{w}}$ that is most similar to the target direction, to show that ICT cannot move efficiently in the target direction because the dimension of the movable latent space is smaller than that of our method. Then, we examined the robustness of each method for the photorealism of the generated images using FID. We also compare the traversal efficiency of the four methods. For the traversal efficiency, we evaluate the cosine similarity between the updated latent code direction and the target direction, the length of a step of latent code traversal, and the cumulative distance traveled in the target direction (Fig. \ref{fig:fid}b). In addition, we further investigate the effect of the scaling factor $\alpha$ on both the robustness and the traversal efficiency of our method. The experimental settings can be found in Supplementary Material Section 2.

\begin{figure}[tb]
  \centering
   \includegraphics[width=1.0\linewidth]{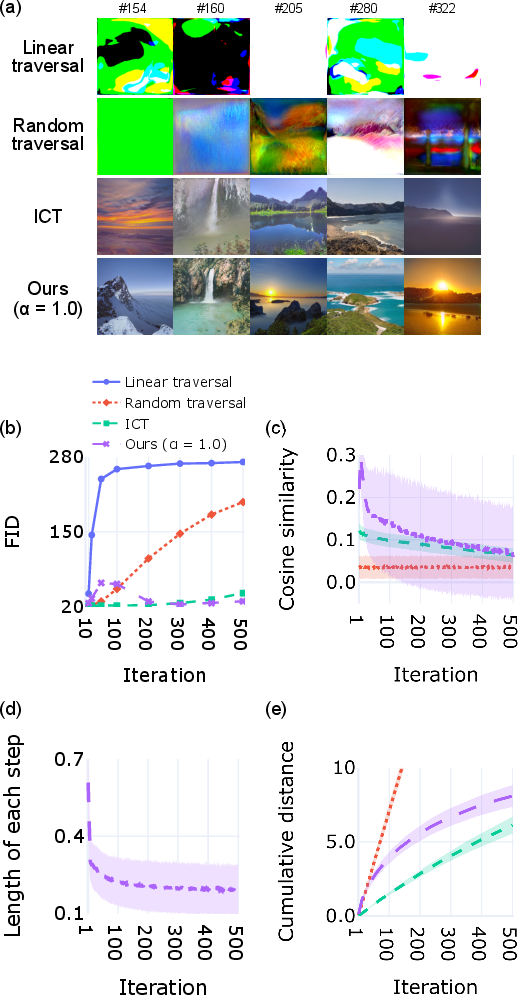}

   \caption{{\bf Evaluation of the Robustness and Traversal Efficiency.} (a) Examples of generated images after 500 iterations. (b) Fr\'{e}chet Inception Distance (FID) scores at each iteration. (c) Cosine similarity between the updated latent code direction and the target direction. (d) Length of each step of latent code traversal in our method. Other methods are omitted for clarity. (e) Cumulative distance traveled in the target direction. The shaded area indicates the $\pm$ standard deviation. Plots for the Linear traversal are omitted in panels (c) and (e) for clarity.}

   \label{fig:stats}
\end{figure}

\noindent {\bf Robustness:} In Fig. \ref{fig:stats}, we show results for StyleGAN2 trained on the LHQ dataset. After 500 iterations, Linear traversal and Random traversal lack photorealism in the generated images, while ICT and our method do not (Fig. \ref{fig:stats}a). Quantitatively, our method has a better (lower) FID score than the other methods after 500 iterations (Fig. \ref{fig:stats}b). Similar results are obtained for other model/dataset combinations (see Supplementary Material Figs. 4, 6, 8, 10, and 12).

\noindent {\bf Traversal Efficiency:} From Fig. \ref{fig:stats}c, it can be seen that the updated latent code direction in our method is more oriented toward the target direction compared with the other methods. In our method, the updated latent code direction faces the target direction at the beginning of the iteration; however, as the iteration progresses, the updated latent code direction gradually turns away from the target direction because of the Bounded Local Space constraint (Fig. \ref{fig:stats}c). The updated latent code direction in ICT is less oriented toward the target direction compared with our method because ICT selects one direction from the Local Basis. From Fig. \ref{fig:stats}d, it can be seen that our method shows dynamic length adjustments at each step as a result of the constraint imposed by Bounded Local Space. With respect to the cumulative distance, our method moves the latent code in the target direction more efficiently than ICT, especially in the early iterations (Fig. \ref{fig:stats}e). To summarize these results, the robustness of our method is comparable to that of ICT, and its traversal efficiency is more efficient than that of ICT, especially in the early iterations. Similar results are obtained for other model/dataset combinations (see Supplementary Material Figs. 4, 6, 8, 10, and 12).

\begin{figure}[tb]
  \centering
   \includegraphics[width=1.0\linewidth]{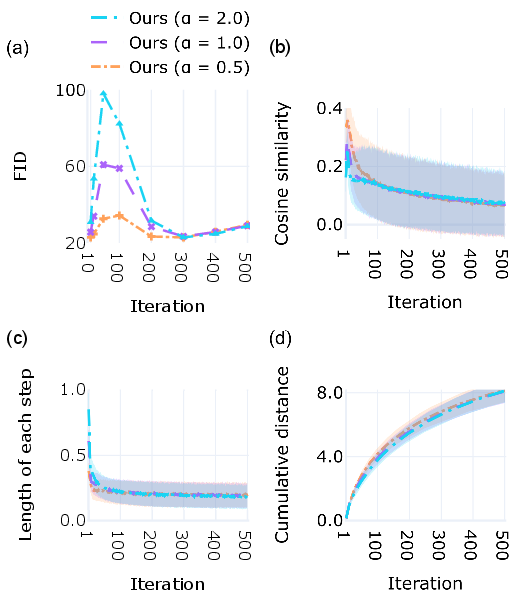}

   \caption{{\bf Evaluation of the Scaling Factor} (a) FID scores at each iteration. (b) Cosine similarity between the updated latent code direction and the target direction. (c) Length of each step of latent code traversal. (d) Cumulative distance traveled in the target direction.}
   \label{fig:alpha}
\end{figure}

\begin{figure*}[t]
  \centering
   \includegraphics[width=0.95\linewidth]{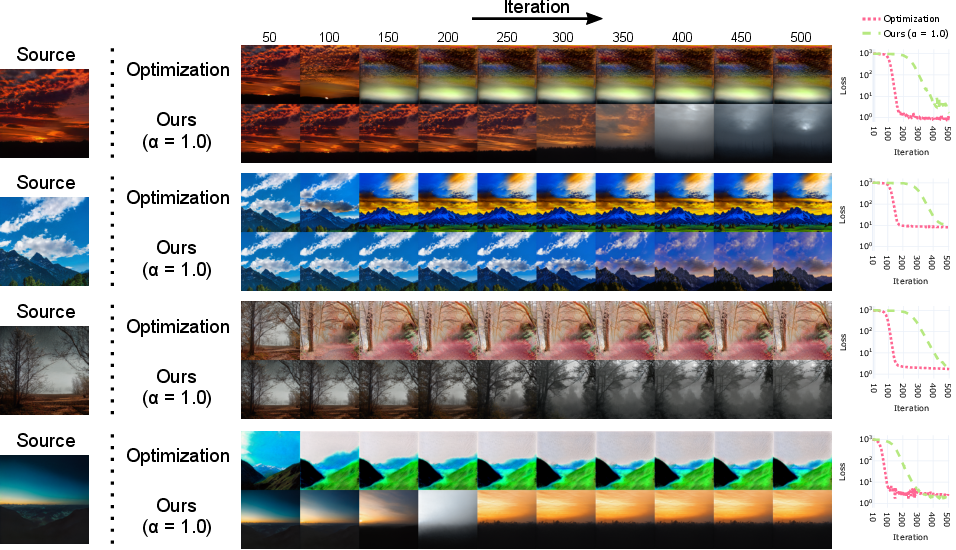}

   \caption{{\bf Aesthetic Manipulation.} Latent codes for source images are manipulated toward the direction that shows a higher aesthetic score for the generated images and away from the initial latent codes. Compared with a baseline method (Optimization), our method (Ours) generates images with maintained photorealism.}
   \label{fig:lapp}
\end{figure*}

\noindent {\bf Scaling Factor:} We investigate the effects of the scaling factor $\alpha$ (Eq. \ref{eq:4}), which determines the size of Bounded Local Space, on the robustness and traversal efficiency of our method. Fig. \ref{fig:alpha} shows the FID score, cosine similarity, length of each step, and cumulative distance for different scaling factors for StyleGAN2 trained on the LHQ dataset. A small scaling-factor setting ($\alpha = 0.5$) reduces the step length, resulting in a lower FID score and larger cosine similarity. This result is because the small step requires many steps before it hits the boundary of the trained latent space. Conversely, a large scaling factor ($\alpha = 2.0$) increases the step length, resulting in a larger FID score and smaller cosine similarity. This result is because the large step can easily exceed the boundary of the trained latent space. 

Interestingly, we find that a large scaling factor worsens the FID score after approximately 50 iterations but that the FID score gradually improves as the number of iterations increases. This result suggests that our method can return from outside the trained latent space because our method attempts to move through the subspace densely mapped by the mapping network. In support of this hypothesis, it is observed that, as the number of iterations increases, colorless or collapsed images return to colorful, realistic, and varied images (see Supplementary Material Figs. 3, 5, and 9). 

Even with different scaling factor magnitudes, the cosine similarity and step lengths are comparable in the early stages of the iterations, and therefore the results at the cumulative length are comparable for different scaling factors. This result means that the latent code we manipulate approaches the boundary of the trained latent space early in the iteration and that our method dynamically controls the step length and the movable subspace. We observed similar results for other model/dataset combinations (see Supplementary Material Figs. 4, 6, 8, 10, and 12). Based on these results, $\alpha = 1.0$ is a simple and effective option regarding a large traversal in a single update and photorealism preservation.

\subsection{Optimized Image Generation}
\label{subsec:app}
We show some applications of our method for different types of latent code optimizations that require large traversals. We compare our method against conventional optimization using stochastic gradient descent. The experiments are performed on randomly sampled latent codes with 500 iterations for optimization. To show the robustness for photorealism of our method, we set a higher learning rate in our method than that used in the conventional optimization. More details and additional results for each experiment can be found in Supplementary Material Section 2.2 and Figs. 14--19.

\begin{figure*}[tb]
  \centering
   \includegraphics[width=0.95\linewidth]{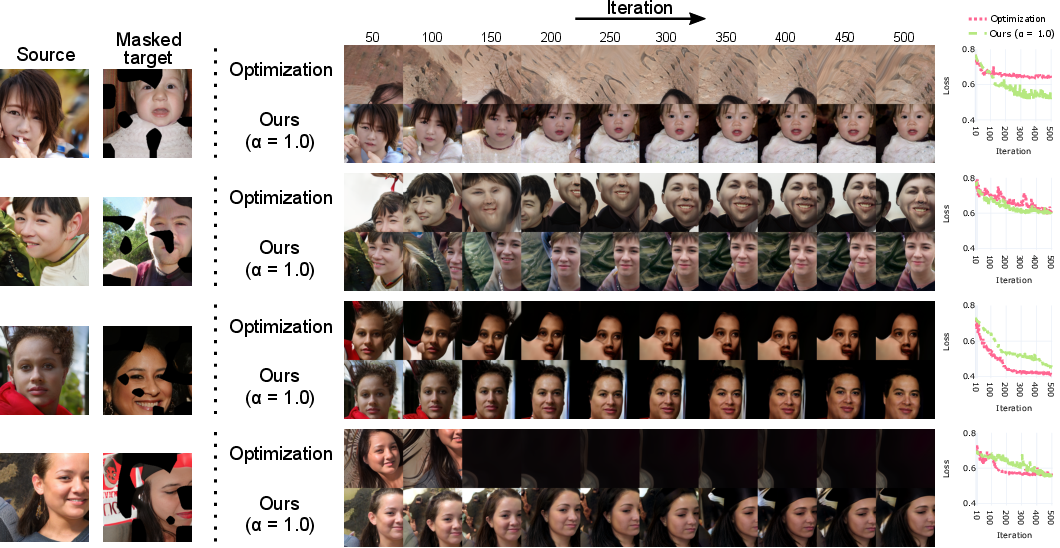}

   \caption{{\bf Latent Search for a Masked Image.} Latent codes for source images are manipulated toward the direction where the generated images are similar to the masked target images. Compared with a baseline method, our method generates images with maintained photorealism.}
   \label{fig:fapp}
\end{figure*}

\begin{figure*}[tb]
  \centering
   \includegraphics[width=0.95\linewidth]{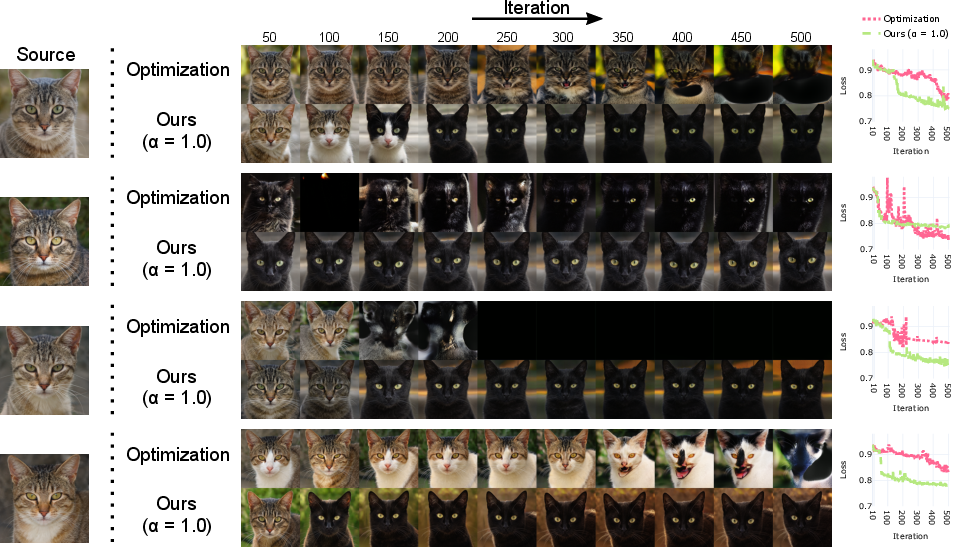}

   \caption{{\bf Text-guided Manipulation.} Latent codes for source images are manipulated toward the direction where the Contrastive Language-Image Pre-training (CLIP) embeddings for the generated images are similar to a CLIP embedding for a text prompt of ``a photo of a black cat''. Compared with the baseline method, our method generates images with maintained photorealism.}
   \label{fig:capp}
\end{figure*}

\noindent {\bf Aesthetic Manipulation:}
Here, we employ an aesthetic score loss $\mathcal{L}_{aes}$ that forces the aesthetic score of the generated image and the target aesthetic score to be the same and a latent space loss $\mathcal{L}_{latent}$ that forces the distance from the initial latent code and the target distance to be the same to search for additional different images from an initial image. The aesthetic score loss is defined as
\begin{equation}
\mathcal{L}_{aes}( \mathbf{x}, s ) = ( E_{aes}(\mathbf{x}) - s )^{2},
\label{eq:11}
\end{equation}
where $\mathbf{x}$ is a generated image, $s$ is a target aesthetic score, and $E_{aes}$ is a prediction network for an aesthetic score of the image. We use Inception-ResNet-v2 \cite{10.5555/3298023.3298188} trained on the AVA dataset \cite{6247954} as the prediction network, which results in a Spearman correlation of 0.71 for the official test dataset. $\mathcal{L}_{latent}$ is defined as
\begin{equation}
\mathcal{L}_{latent}( \mathbf{w}, \mathbf{w}_{0}, d_{t} ) = (  \| \mathbf{w} - \mathbf{w}_{0}  \|^{2}_{2} - d_t )^{2},
\label{eq:12}
\end{equation}
where $\mathbf{w}$ is a manipulating latent code, $\mathbf{w}_{0}$ is an initial latent code, and $d_t$ is a target distance.
We use StyleGAN2 pretrained on the LHQ dataset. We set $s$ and $d_t$ to 8.51, which is the maximum predicted aesthetic score in the LHQ dataset, and 100, respectively. We set the learning rate to 1 $\times 10^{-3}$ for conventional optimization and 5 $\times 10^{-3}$ for our method.
Fig. \ref{fig:lapp} shows the optimization result. In conventional optimization, the realism of the generated image is gradually lost; however, our method maintains realism.

\noindent {\bf Latent Search for a Masked Image:}
Here, we employ a loss using the perceptually based pairwise image distance \cite{Karras_2019_CVPR} between a generated image and a target image that forces the generated image and the target image to be the same. The target image is set to a randomly masked image. We use StyleGAN3 pretrained on the FFHQ dataset and set the learning rate to 5.0 for the conventional optimization and 10 for our method. From Fig. \ref{fig:fapp}, it can be seen that our method achieves good performance for photorealism.

\noindent {\bf Text-guided Manipulation:}
Here, we use the Contrastive Language-Image Pre-training (CLIP) \cite{pmlr-v139-radford21a} model, which learns a text-image embedding space, and employs a CLIP loss $\mathcal{L}_{CLIP}$, which forces the embeddings of a generated image and those of a text prompt to be the same.
\begin{equation}
\mathcal{L}_{CLIP} ( \mathbf{w}, t ) = D_{sph} ( E_{I}(G(\mathbf{w})), E_{t}(t)  )
\label{eq:13}
\end{equation}
Here, $\mathbf{w}$ is a latent code to be manipulated, $t$ is a text prompt, $D_{sph}$ is the spherical distance, $E_{I}$ is a CLIP image encoder, $G$ is a synthesis network of EG3D, and $E_{t}$ is a CLIP text encoder. The text prompt is set to be ``a photo of a black cat''. We use EG3D pretrained on the AFHQv2 Cats dataset and set the learning rate to 1.0 for the conventional optimization and 50 for our method. Fig. \ref{fig:capp} shows that our method displays black cat images that do not lack realism.

\section{Limitations}
\label{sec:lim}

\noindent {\bf Applicable Latent Space:}
Our method requires tracking the latent codes in the $\mathcal{Z}$ and $\mathcal{W}$ spaces for successive latent manipulating because our method requires the Jacobian matrix of a mapping network. This requirement limits some applications that cannot obtain these latent codes; for example, recent methods for GAN inversion \cite{Richardson_2021_CVPR} only return latent codes in the $\mathcal{W}^{+}$ space, which is an extended latent space of $\mathcal{W}$.

\noindent {\bf Unrealistic images in in-distribution:}
Our method is not guaranteed to provide photo-realistic images when the style-based model generates collapsed images from latent codes sampled from the training distribution in $\mathcal{Z}$ space because the performance of our method depends on a trained mapping network. In fact, we do observe some unrealistic images when using our method.

\section{Conclusions}
\label{sec:con}
We present a latent traversal algorithm within Bounded Local Space, enabling significant latent manipulation while maintaining photorealism. By leveraging a mapping network of the style-based model, we can traverse a trained distribution in the latent space. Extensive experiments indicate that our method can be applied to different types of optimizations. We believe that our method will enable many applications that require broad explorations in the latent space.

{\small
\bibliographystyle{ieee_fullname}
\bibliography{egbib2}
}

\end{document}


\title{Trained Latent Space Navigation to Prevent Lack of Photorealism in Generated Images on Style-based Models\\ ${}$ \\ ${}$ Supplementary Material}

\author{Takumi Harada${}^{1}$, Kazuyuki Aihara${}^{2}$, Hiroyuki Sakai${}^{1}$\\
${}^{1}$TOYOTA CENTRAL R\&D LABS., INC.   ${}^{2}$University of Tokyo\\
{\tt\small t-harada@mosk.tytlabs.co.jp, aihara@u-tokyo.ac.jp, sakai@mosk.tytlabs.co.jp}}

\maketitle

\section{Implementation Details}
\label{sec:imp}

All experiments are conducted using Pytorch 1.10.1 and are run on an NVIDIA A100 GPU and an AMD EPYC 7742. \\

\noindent {\bf Our Method:} To prevent moving latent codes in unreliable directions and to avoid effects of numerical instability for singular value decomposition \cite{torchsvd}, elements of the coefficient vector $\mathbf{A}$ corresponding to singular vectors with singular values of 0.05 or less are set to 0.\\

\noindent {\bf StyleGAN2:} For the landscape dataset, we use a StyleGAN2 \cite{Karras_2020_CVPR} model trained on the LHQ dataset \cite{Skorokhodov_2021_ICCV} with $256 \times 256$ resolution, 155.2M trained images shown to the discriminator, and the default settings in \cite{stylegan2ada}. The LHQ dataset is published under the CC BY 2.0 license (details can be found in \cite{lhqc}). For the butterfly dataset \cite{butterfly}, we use a StyleGAN2 model trained with $256 \times 256$ resolution, 24.8M trained images shown to the discriminator, and the default settings in \cite{stylegan2ada}. The images in the butterfly dataset are provided by the Natural History Museum under the CC BY 4.0 license, and code for the dataset are provided under the MIT license \cite{butterfly}. The codes for StyleGAN2 are available at \cite{stylegan2ada} under the Nvidia Source Code License for StyleGAN2 with ADA \cite{nlc3}. \\

\noindent {\bf StyleGAN3:} For the face dataset, we use a StyleGAN3 \cite{NEURIPS2021_076ccd93} pretrained model for config T on the FFHQ dataset \cite{Karras_2019_CVPR} with $256 \times 256$ resolution. For the WikiArt dataset \cite{wikiart}, we use a StyleGAN3 pretrained model for config T with $1024 \times 1024$ resolution (model weights are available at \cite{wikiart3}). The WikiArt dataset can be used under the terms and conditions of WikiArt.org \cite{wikitc}. The codes for StyleGAN3 are available at \cite{stylegan3} under the Nvidia Source Code License for StyleGAN3 \cite{nlc3}.\\

\noindent {\bf SemanticStyleGAN:} We use a SemanticStyleGAN \cite{Shi_2022_CVPR} model pretrained on the CelebAMask-HQ dataset \cite{CelebAMask-HQ} with $512 \times 512$ resolution (model weights are available at \cite{ssg}). The CelebAMask-HQ dataset is available for non-commercial research and educational purposes \cite{celeb}. The codes for SemanticStyleGAN are available under the CC BY-NC-SA 4.0 license \cite{ssg}.\\

\noindent {\bf EG3D:} We use a EG3D \cite{Chan_2022_CVPR} model pretrained on the AFHQv2 Cats dataset \cite{9157662, NEURIPS2021_076ccd93} with $512 \times 512$ resolution (model weights are available at \cite{eg3dgit}). We set the truncation psi to 0.5, truncation cutoff to 8, azimuthal angle to $\pi /2$, polar angle $\pi /2 - 0.2$, and the default settings in \cite{eg3dgit}. The AFHQv2 Cats dataset is provided under the CC BY-NC 4.0 license \cite{catlc}. The codes for EG3D are available under the NVIDIA Source Code License for EG3D \cite{eg3dlc}.

\section{Experiments}
\label{sec:exp}

\subsection{Latent Traversal}
\label{subsec:lat}

In the latent traversal experiment, the latent code is moved for 500 iterations, with the distance traveled in each iteration being a specific Euclidean distance. To check the performance of our method under various conditions, we set various distances traveled in each iteration: 2.0 for StyleGAN2 trained on the LHQ dataset, 10 for StyleGAN2 trained on the butterfly dataset, 20 for StyleGAN3 trained on the FFHQ and WikiArt datasets, 5.0 for SemanticStyleGAN, and 20 for EG3D. The distances are much larger than in previous latent code manipulations for attribute editing; for example, 0.02--0.16 step size in \cite{choi2022do}, 3.0 as a total traversal distance in \cite{Shen_2020_CVPR}, and 0.2 step-size coefficient between initial attributes and target attributes in \cite{Li_2021_CVPR}.\\

\begin{figure*}[p]
  \centering
   \includegraphics[width=0.8\linewidth]{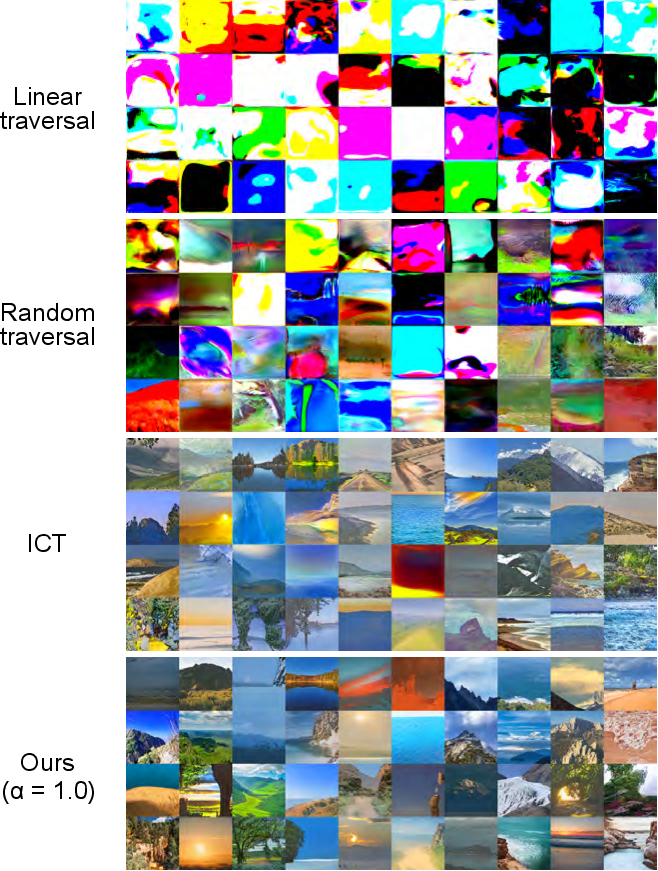}
   
   \caption{ {\bf Latent Traversal Experiment for StyleGAN2 Trained on the LHQ Dataset.} Randomly selected examples of generated images after 500 iterations in each method. }
   \label{fig:exalan}
\end{figure*}

\begin{figure*}[p]
  \centering
   \includegraphics[width=1.0\linewidth]{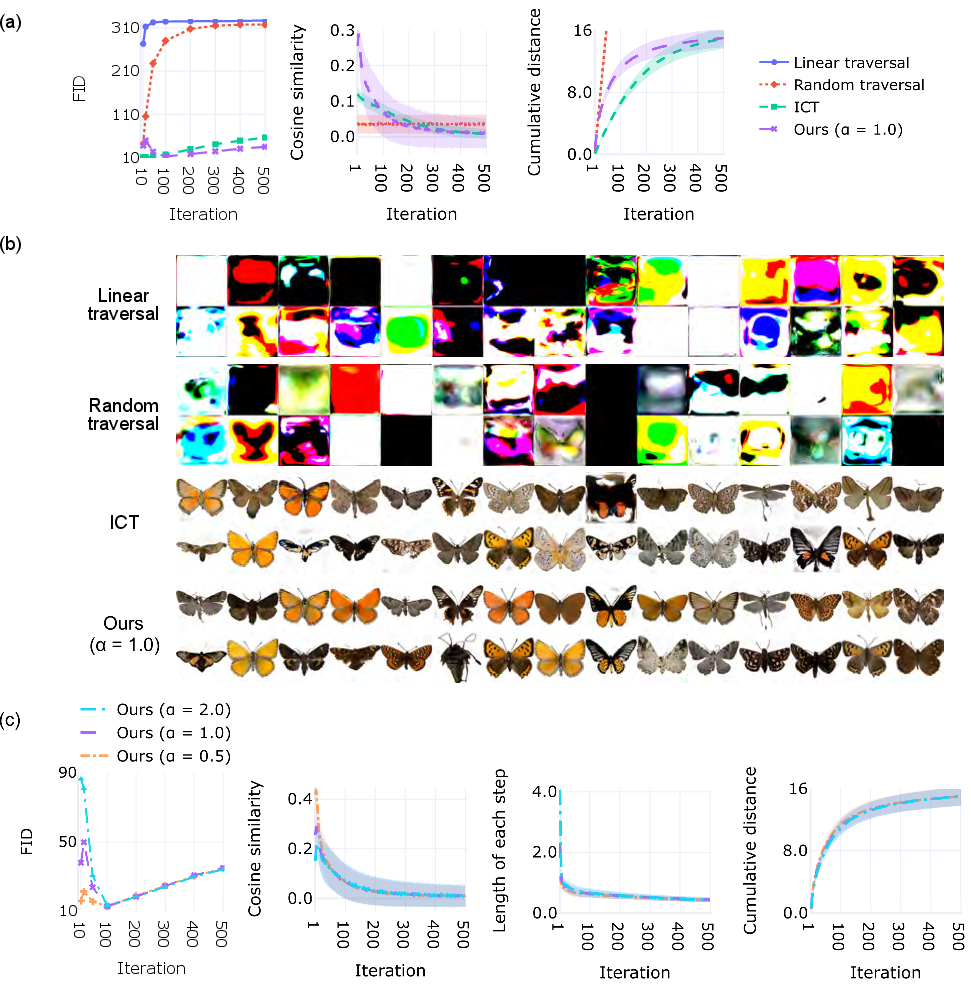}
   
   \caption{ {\bf Latent Traversal Experiment for StyleGAN2 Trained on the Butterfly Dataset.} (a) Fr\'{e}chet Inception Distance (FID) scores at each iteration (left), cosine similarity between the updated latent code direction and the target direction (middle), and cumulative distance traveled in the target direction (right). (b) Randomly selected generated images after 500 iterations in each method. (c) Evaluation of the scaling factor. From left to right, FID scores at each iteration, cosine similarity between the updated latent code direction and the target direction, length of each step of latent code traversal, and cumulative distance traveled in the target direction. }
   \label{fig:statbut}
\end{figure*}

\begin{figure*}[p]
  \centering
   \includegraphics[width=1.0\linewidth]{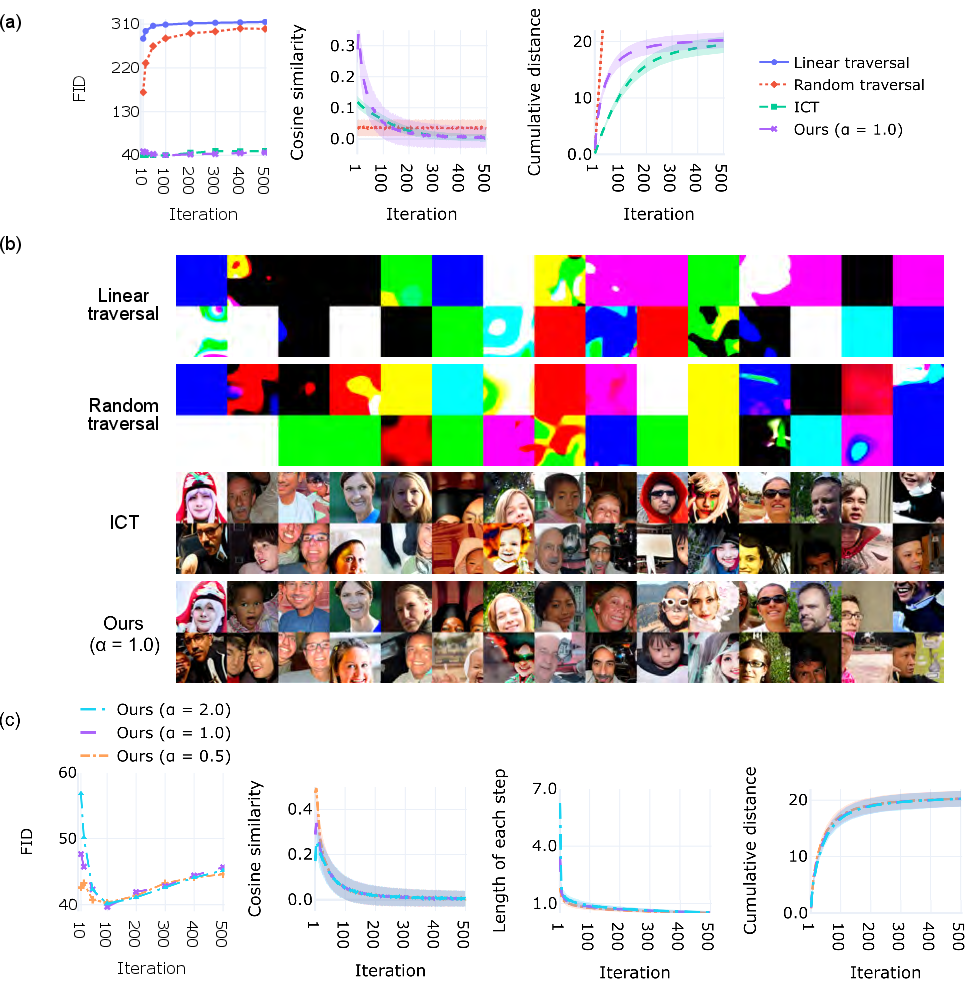}
   
   \caption{  {\bf Latent Traversal Experiment for StyleGAN3 Trained on the FFHQ Dataset.} (a) FID scores at each iteration (left), cosine similarity between the updated latent code direction and the target direction (middle), and cumulative distance traveled in the target direction (right). (b) Randomly selected generated images after 500 iterations in each method. (c) Evaluation of the scaling factor. From left to right, FID scores at each iteration, cosine similarity between the updated latent code direction and the target direction, length of each step of latent code traversal, and cumulative distance traveled in the target direction. }
   \label{fig:statfac}
\end{figure*}

\begin{figure*}[p]
  \centering
   \includegraphics[width=1.0\linewidth]{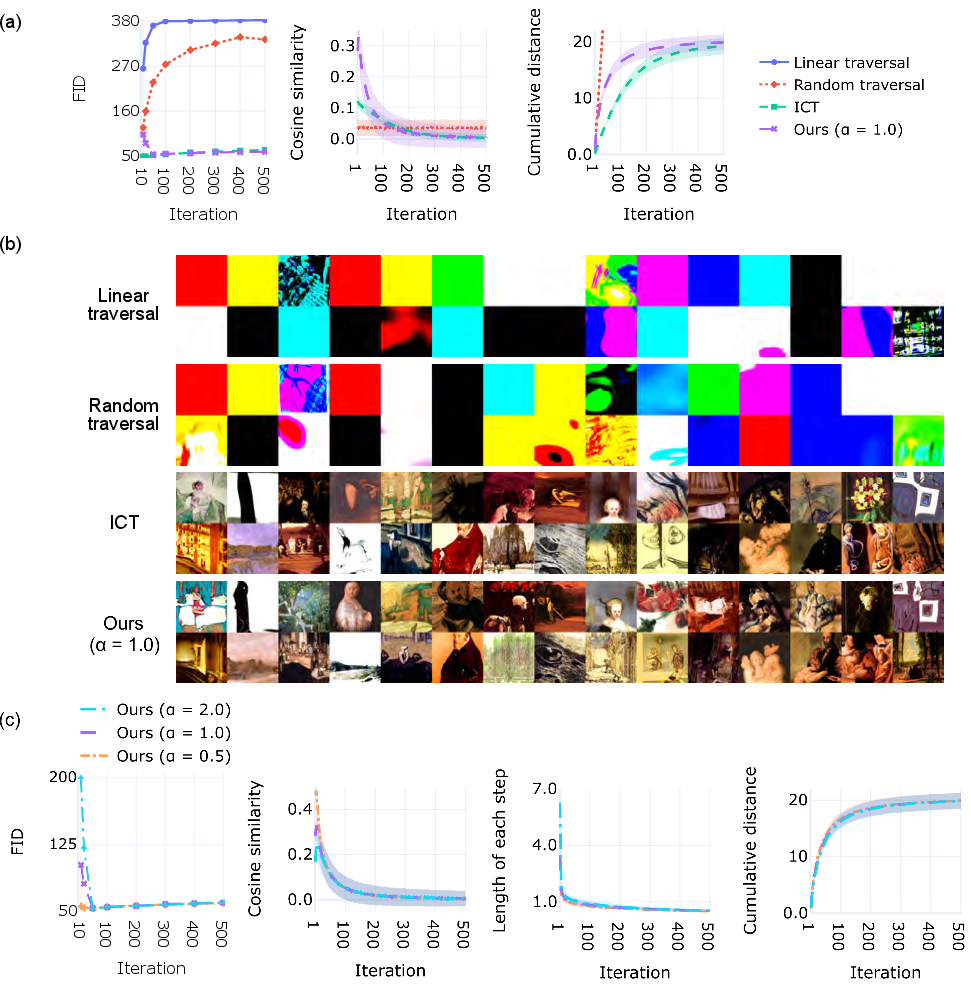}
   
   \caption{ {\bf Latent Traversal Experiment for StyleGAN3 Trained on the WikiArt Dataset.} (a) FID scores at each iteration (left), cosine similarity between the updated latent code direction and the target direction (middle), and cumulative distance traveled in the target direction (right). (b) Randomly selected generated images after 500 iterations in each method. (c) Evaluation of the scaling factor. From left to right, FID scores at each iteration, cosine similarity between the updated latent code direction and the target direction, length of each step of latent code traversal, and cumulative distance traveled in the target direction. }
   \label{fig:statwik}
\end{figure*}

\begin{figure*}[p]
  \centering
   \includegraphics[width=1.0\linewidth]{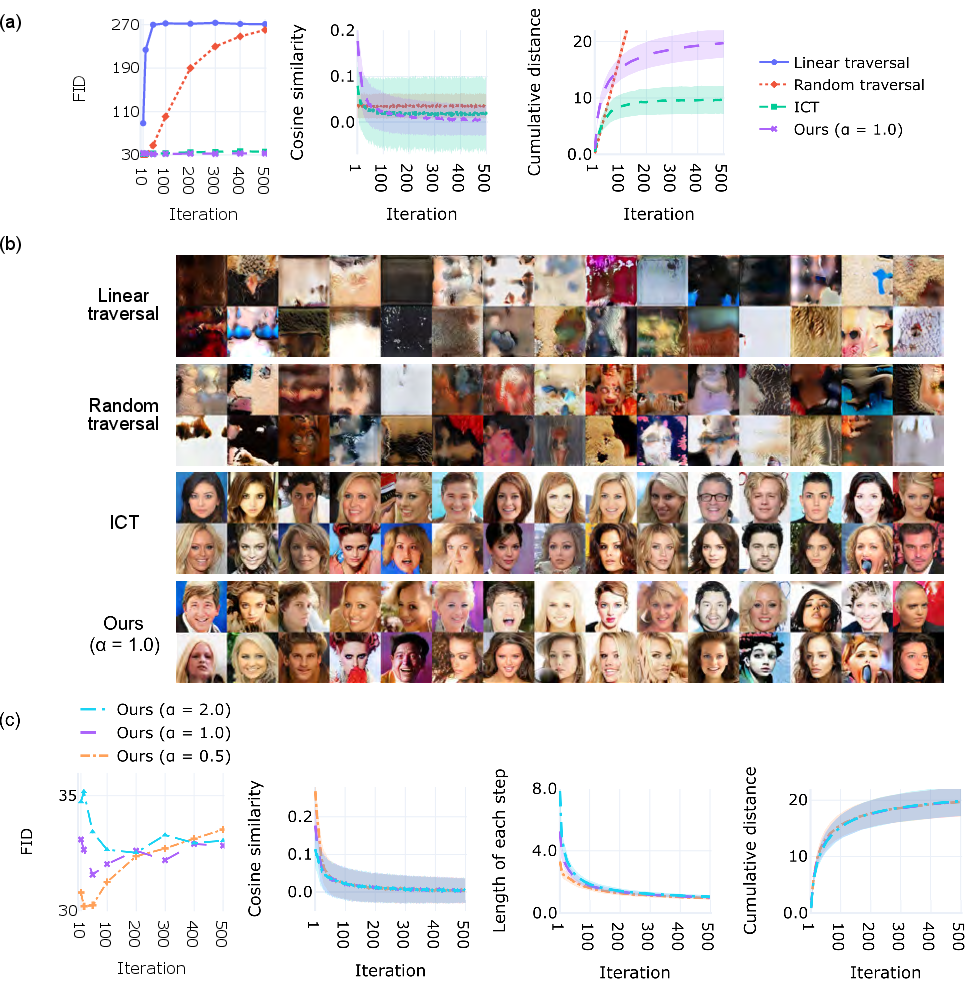}
   
   \caption{ {\bf Latent Traversal Experiment for SemanticStyleGAN Trained on the CelebAMask-HQ Dataset.} (a) FID scores at each iteration (left), cosine similarity between the updated latent code direction and the target direction (middle), and cumulative distance traveled in the target direction (right). (b) Randomly selected generated images after 500 iterations in each method. (c) Evaluation of the scaling factor. From left to right, FID scores at each iteration, cosine similarity between the updated latent code direction and the target direction, length of each step of latent code traversal, and cumulative distance traveled in the target direction.}
   \label{fig:statsem}
\end{figure*}

\begin{figure*}[p]
  \centering
   \includegraphics[width=1.0\linewidth]{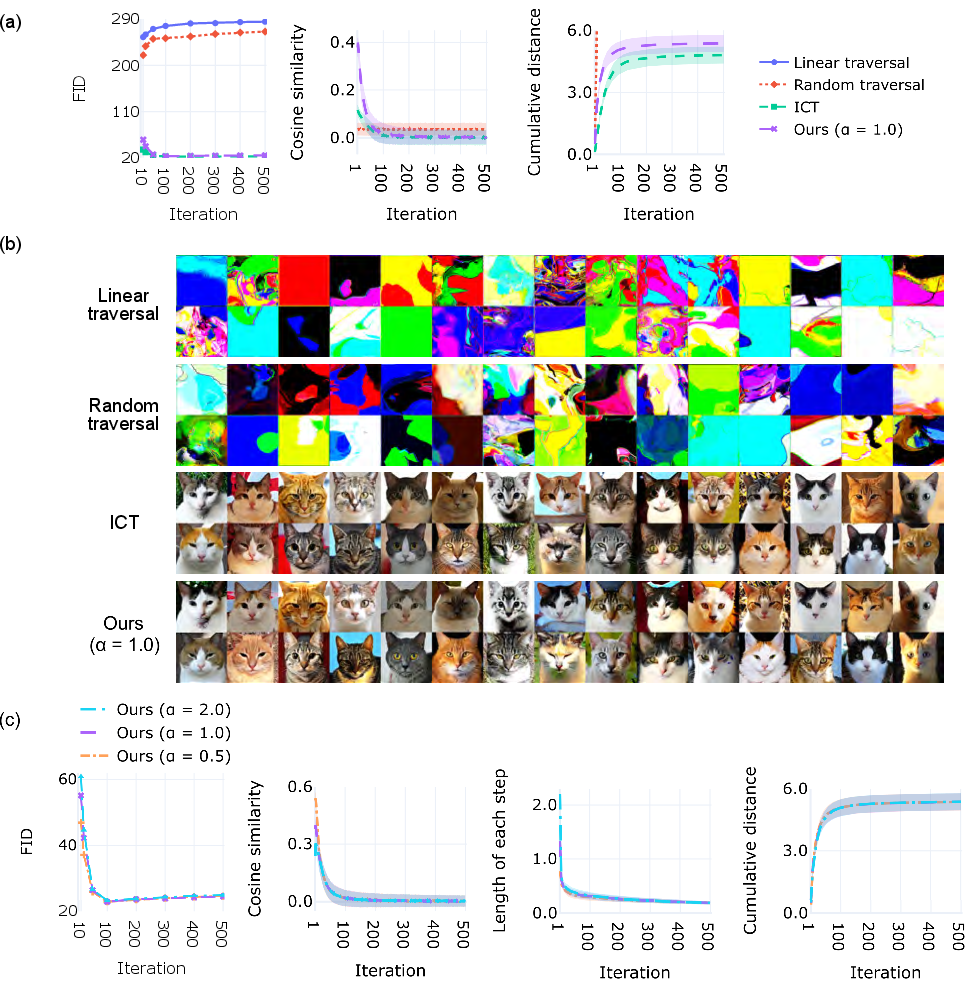}
   
   \caption{ {\bf Latent Traversal Experiment for EG3D Trained on the AFHQv2 Cats Dataset.} (a) FID scores at each iteration (left), cosine similarity between the updated latent code direction and the target direction (middle), and cumulative distance traveled in the target direction (right). (b) Randomly selected generated images after 500 iterations in each method. (c) Evaluation of the scaling factor. From left to right, FID scores at each iteration, cosine similarity between the updated latent code direction and the target direction, length of each step of latent code traversal, and cumulative distance traveled in the target direction.}
   \label{fig:stateg3}
\end{figure*}

\noindent {\bf Robustness:} We evaluate the robustness of photorealism (main document Section 4.1). More results of generated images in StyleGAN2 trained on the LHQ dataset are shown in Fig. \ref{fig:exalan}. We also show the results for StyleGAN2 trained on the butterfly dataset (Fig. \ref{fig:statbut}b, Fr\'{e}chet Inception Distance (FID) scores in Fig. \ref{fig:statbut}a), StyleGAN3 trained on the FFHQ dataset (Fig. \ref{fig:statfac}b, FID scores in Fig. \ref{fig:statfac}a), StyleGAN3 trained on the WikiArt dataset (Fig. \ref{fig:statwik}b, FID scores in Fig. \ref{fig:statwik}a), SemanticStyleGAN (Fig. \ref{fig:statsem}b, FID scores in Fig. \ref{fig:statsem}a), and EG3D (Fig. \ref{fig:stateg3}b, FID scores in Fig. \ref{fig:stateg3}a).\\

\begin{figure}[tb]
  \centering
   \includegraphics[width=1.0\linewidth]{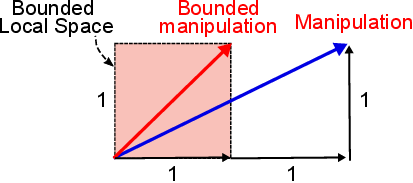}
   
   \caption{ {\bf Abstract Example for the Updated Latent Code Direction in Our Method.} Here is an example where the target latent code advances 2 horizontally and 1 vertically, and the bounded local space ranges from 1 horizontally to 1 vertically. Our method manipulates the latent codes within Bounded Local Space. Therefore, the manipulated latent code is placed at the closest point to the target latent code within Bounded Local Space.}
   \label{fig:cos}
\end{figure}

\noindent {\bf Traversal Efficiency:} To examine the traversal efficiency in each method, we evaluate the cosine similarity between the updated latent code direction and the target direction (main document Section 4.1). Note that the updated latent code direction in our method does not necessarily face the target direction (see Fig. \ref{fig:cos}). For ICT and our method, we measure the cosine similarity, the length of a step, and the cumulative distance using $\Delta \mathbf{w} = \mathbf{w}_{t+1} - \mathbf{w}_{t}$, where $\mathbf{w}_{t} = M(\mathbf{z}_{t})$ and $\mathbf{w}_{t+1} = M(\mathbf{z}_{t+1})$, after calculating $\mathbf{z}_{t+1}$ from $\mathbf{z}_{t}$.

We show the results of the cosine similarity and the cumulative distance for StyleGAN2 trained on the butterfly dataset (Fig. \ref{fig:statbut}a), StyleGAN3 trained on the FFHQ dataset (Fig. \ref{fig:statfac}a), StyleGAN3 trained on the WikiArt dataset (Fig. \ref{fig:statwik}a), SemanticStyleGAN (Fig. \ref{fig:statsem}a), and EG3D (Fig. \ref{fig:stateg3}a).\\

\begin{figure*}[p]
  \centering
   \includegraphics[width=1.0\linewidth]{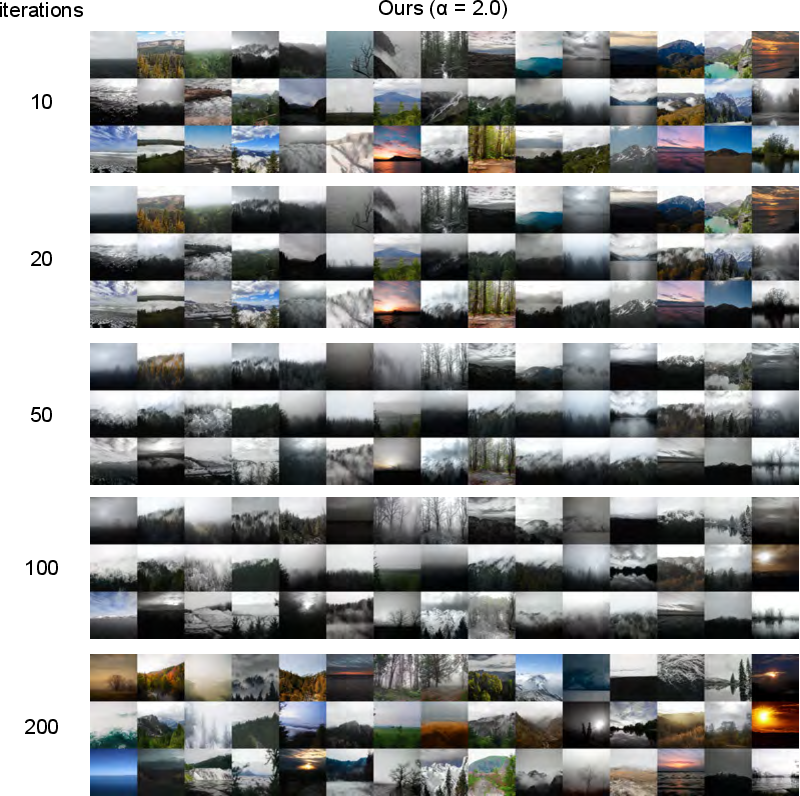}
   
   \caption{ {\bf Latent Traversal Experiment for StyleGAN2 Trained on the LHQ Dataset.} Randomly selected examples of generated images at each iteration in our method using the scaling factor $\alpha = 2.0$.}
   \label{fig:exalans}
\end{figure*}

\begin{figure*}[p]
  \centering
   \includegraphics[width=1.0\linewidth]{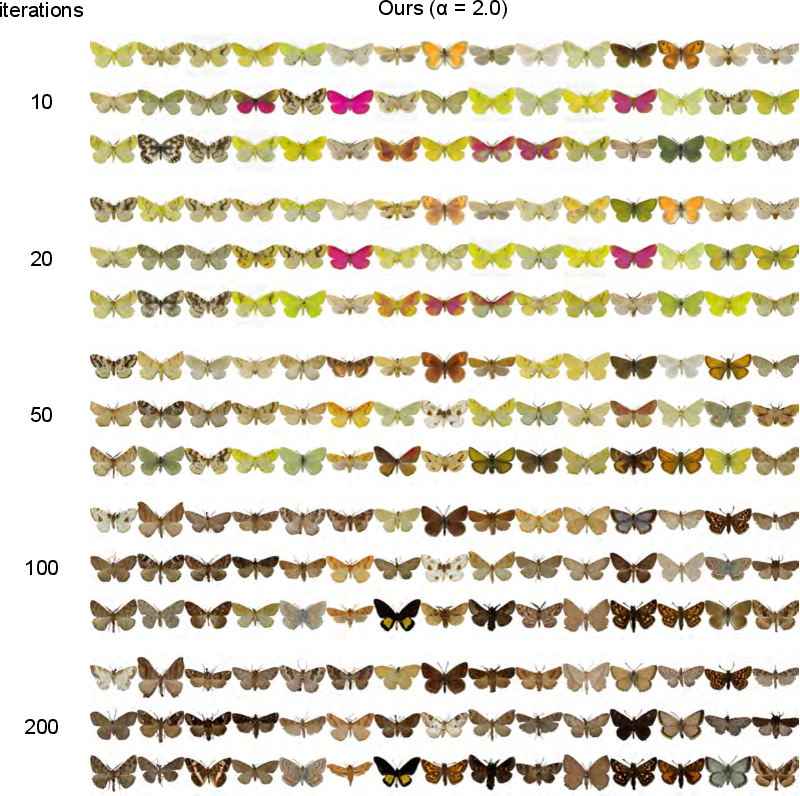}
   
   \caption{ {\bf Latent Traversal Experiment for StyleGAN2 Trained on the Butterfly Dataset.} Randomly selected examples of generated images at each iteration in our method using the scaling factor $\alpha = 2.0$.}
   \label{fig:statbuts}
\end{figure*}

\begin{figure*}[p]
  \centering
   \includegraphics[width=1.0\linewidth]{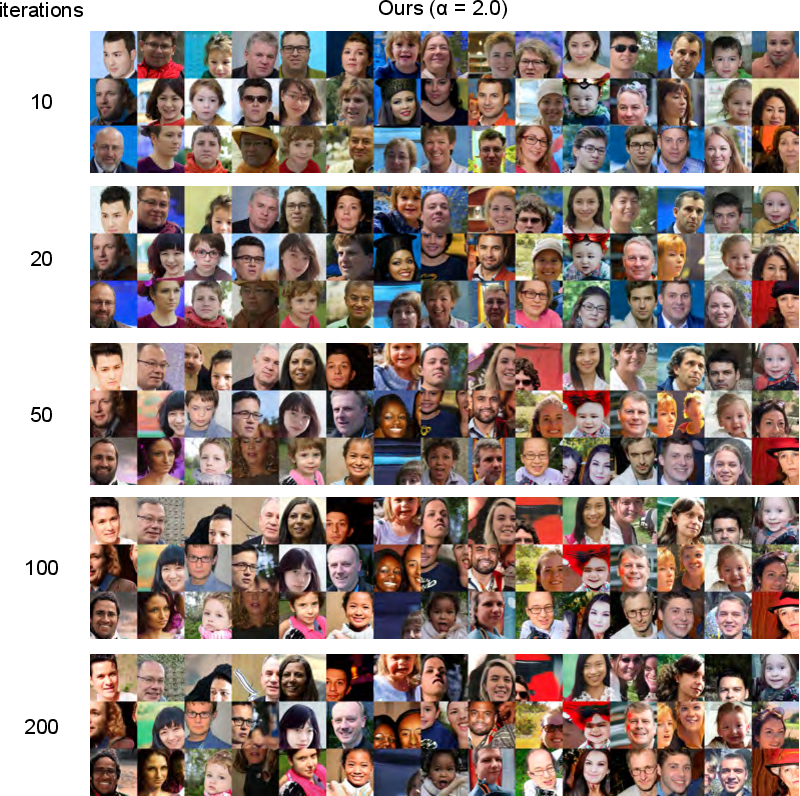}
   
   \caption{{\bf Latent Traversal Experiment for StyleGAN3 Trained on the FFHQ Dataset.} Randomly selected examples of generated images at each iteration in our method using the scaling factor $\alpha = 2.0$.}
   \label{fig:statcs}
\end{figure*}

\begin{figure*}[p]
  \centering
   \includegraphics[width=1.0\linewidth]{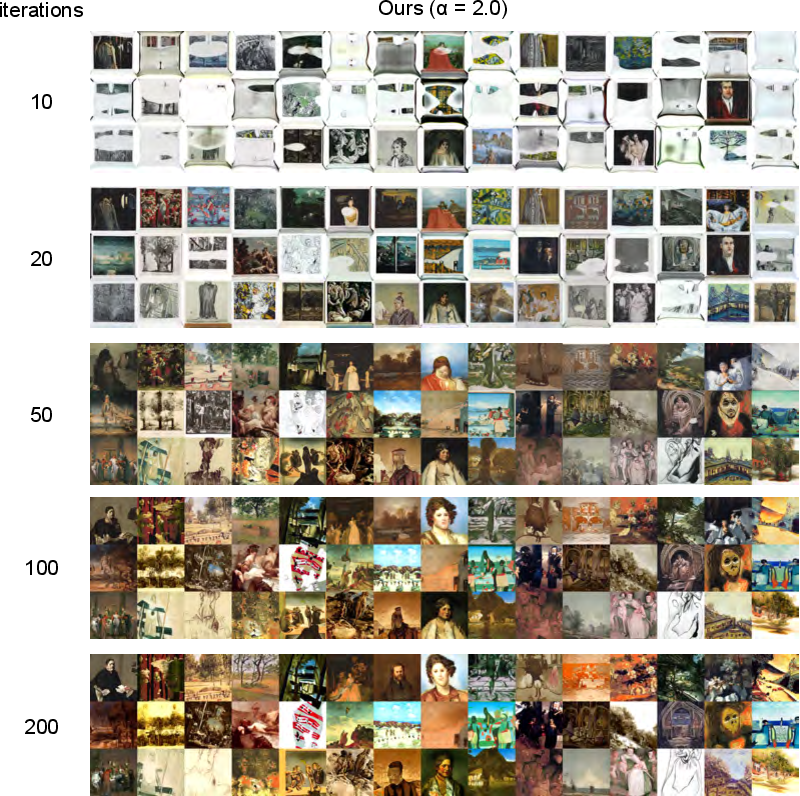}
   
   \caption{ {\bf Latent Traversal Experiment for StyleGAN3 Trained on the WikiArt Dataset.} Randomly selected examples of generated images at each iteration in our method using the scaling factor $\alpha = 2.0$. }
   \label{fig:statwiks}
\end{figure*}

\begin{figure*}[p]
  \centering
   \includegraphics[width=1.0\linewidth]{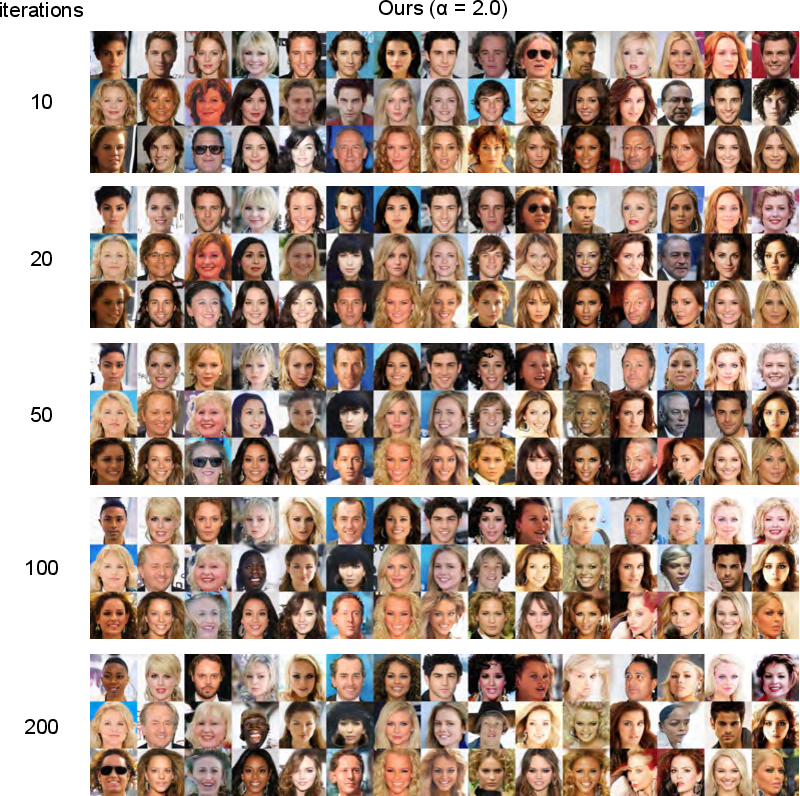}
   
   \caption{ {\bf Latent Traversal Experiment for SemanticStyleGAN Trained on the CelebAMask-HQ Dataset.} Randomly selected examples of generated images at each iteration in our method using the scaling factor $\alpha = 2.0$.}
   \label{fig:statsems}
\end{figure*}

\begin{figure*}[p]
  \centering
   \includegraphics[width=1.0\linewidth]{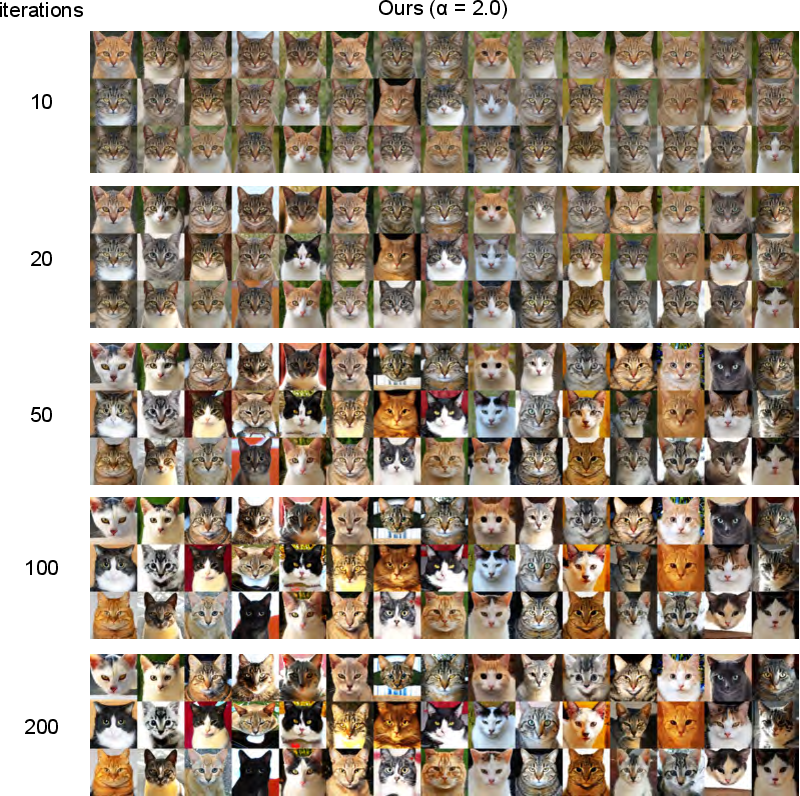}
   
   \caption{ {\bf Latent Traversal Experiment for EG3D Trained on the AFHQv2 Cats Dataset.} Randomly selected examples of generated images at each iteration in our method using the scaling factor $\alpha = 2.0$.}
   \label{fig:stateg3s}
\end{figure*}

\noindent {\bf Scaling Factor:} We investigate the effects of the scaling factor $\alpha$, which determines the size of Bounded Local Space in our method (main document Section 4.1). We show the results for StyleGAN2 trained on the butterfly dataset (Fig. \ref{fig:statbut}c), StyleGAN3 trained on the FFHQ dataset (Fig. \ref{fig:statfac}c), StyleGAN3 trained on the WikiArt dataset (Fig. \ref{fig:statwik}c), SemanticStyleGAN (Fig. \ref{fig:statsem}c), and EG3D (Fig. \ref{fig:stateg3}c). Randomly selected examples of generated images for our method using the scaling factor $\alpha = 2.0$ are shown (StyleGAN2 trained on the LHQ dataset in Fig. \ref{fig:exalans}; StyleGAN2 trained on the butterfly dataset in Fig. \ref{fig:statbuts}; StyleGAN3 trained on the FFHQ dataset in Fig. \ref{fig:statcs}; StyleGAN3 trained on the WikiArt dataset in Fig. \ref{fig:statwiks}; SemanticStyleGAN in Fig. \ref{fig:statsems}; and EG3D in Fig. \ref{fig:stateg3s}). A large scaling factor ($\alpha = 2.0$) results in high FID scores at the beginning of the iteration; however, the FID scores gradually decrease with the increasing number of iterations. Upon reviewing the generated images, we find that a high FID score results in hazy and colorless images for StyleGAN2 trained on the LHQ dataset (50 iterations in Fig. \ref{fig:exalans}), images with little variation for StyleGAN2 trained on the butterfly dataset (10 iterations in Fig. \ref{fig:statbuts}), and collapsed images for StyleGAN3 trained on the WikiArt dataset (10 iterations in Fig. \ref{fig:statwiks}). As the iterations progress, the images become more colorful, realistic, and varied. Therefore, the results suggest that, in our method, even if the latent code deviates to out-of-distribution, it will return to in-distribution after repeated iterations because the latent code moves through the region that the mapping network densely maps. \\

\noindent {\bf Computation Time:}
The average time per input for obtaining the Jacobian matrix for 1,000 random inputs is $5.32\times10^{-3}$, $7.74\times10^{-3}$, and $2.74\times10^{-1}$ s/input in StyleGAN2, StyleGAN3, and EG3D, respectively. The average time per input for computing SVD of the Jacobian matrix for 1,000 random inputs is $8.97\times10^{-2}$, $7.70\times10^{-2}$, and $5.93\times10^{-2}$ s/input in StyleGAN2, StyleGAN3, and EG3D, respectively.

\subsection{Optimized Image Generation}
\label{subsec:app}
\begin{figure*}[p]
  \centering
   \includegraphics[width=1.0\linewidth]{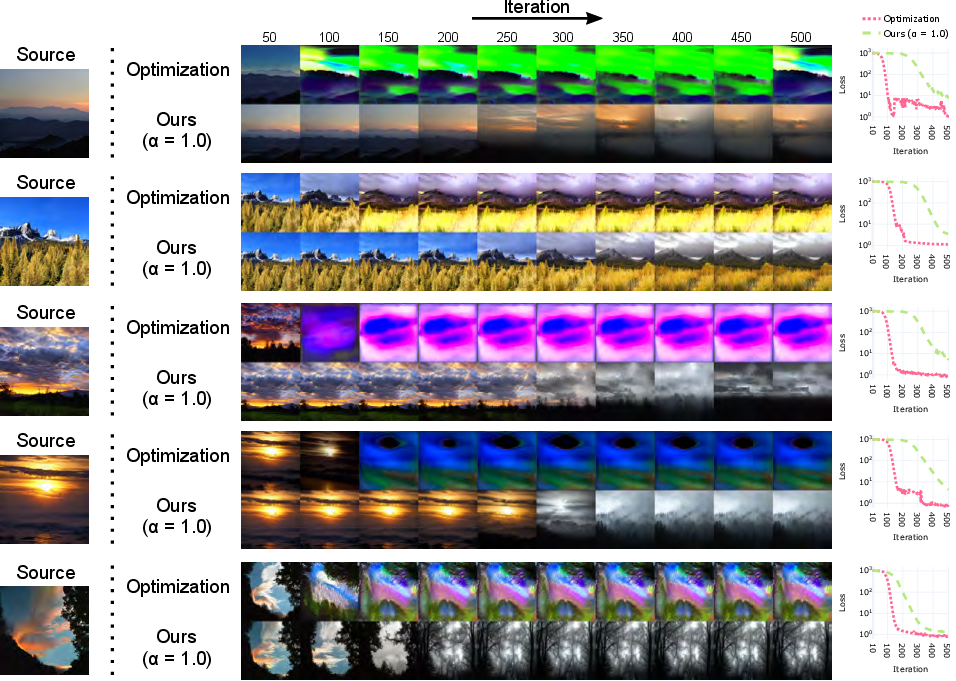}
   
   \caption{ {\bf Aesthetic Manipulation.} Latent codes for source images are manipulated toward the direction that shows a higher aesthetic score for the generated images and away from the initial latent codes. Compared with the baseline method (Optimization), our method (Ours) results in generated images that maintain photorealism.}
   \label{fig:laapp}
\end{figure*}

\begin{figure*}[p]
  \centering
   \includegraphics[width=1.0\linewidth]{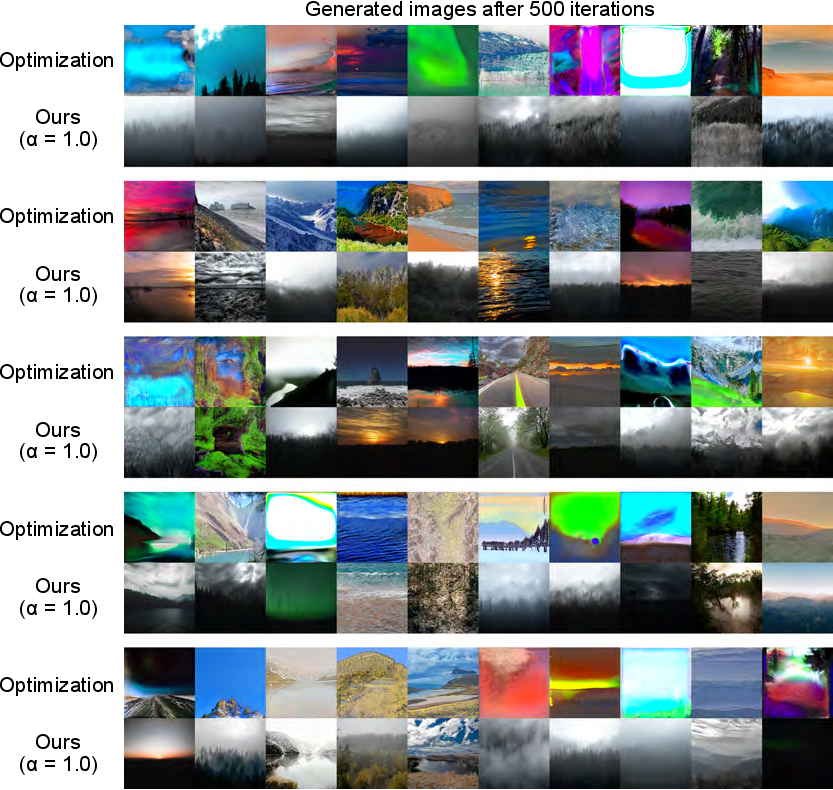}
   
   \caption{{\bf Aesthetic Manipulation.} Randomly selected examples of generated images after 500 iterations.}
   \label{fig:laapp2}
\end{figure*}

\noindent {\bf Aesthetic Manipulation:} 
Here, we use Inception-ResNet-v2 \cite{10.5555/3298023.3298188} trained on the AVA dataset \cite{6247954} as the prediction network, which resulted in a Pearson correlation of 0.71, a Spearman correlation of 0.71, and a Concordance correlation of 0.67 for the official test dataset. This model is fine-tuned from a pretrained model on the ImageNet dataset \cite{ILSVRC15} following an official train--test data split for 100 epochs using early stopping with a patience of 10 and a batch size of 32. We use the SGD optimizer with a learning rate of 1 $\times 10^{-3}$ and a momentum of 0.9 using a decaying learning rate of 0.1 at every 10 epochs. The prediction accuracy is evaluated using the correlation between the true and predicted values. Additional results for Section 4.2 in the main document are shown in Fig. \ref{fig:laapp}. More results of randomly selected generated images after 500 iterations are shown in Fig. \ref{fig:laapp2}.\\

\begin{figure*}[p]
  \centering
   \includegraphics[width=1.0\linewidth]{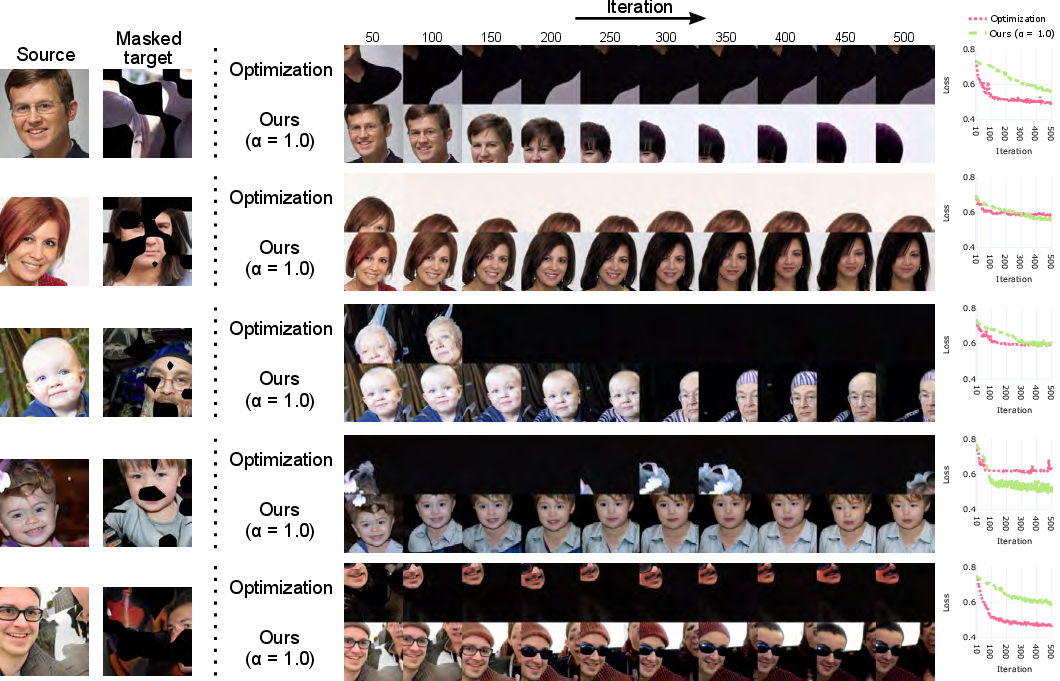}

   \caption{ {\bf Latent Search for a Masked Image.} Latent codes for source images are manipulated toward the direction where generated images are similar to the masked target image. Compared with the baseline method, our method results in generated images that maintain photorealism. }
   \label{fig:fapp}
\end{figure*}

\begin{figure*}[p]
  \centering
   \includegraphics[width=1.0\linewidth]{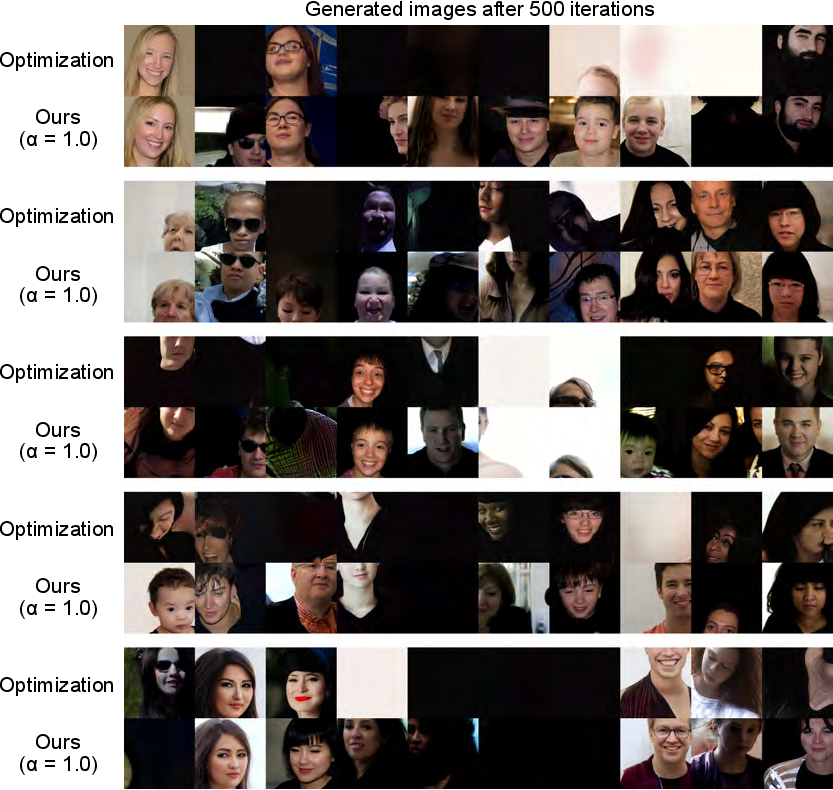}

   \caption{ {\bf Latent Search for a Masked Image.} Randomly selected examples of generated images after 500 iterations.}
   \label{fig:fapp2}
\end{figure*}

\noindent {\bf Latent Search for a Masked Image:} 
The target masked image is made via the following procedure. First, a random mask is created with $6 \times 6$ resolution and is upsampled to $256 \times 256$ resolution using bilinear interpolation. The mask is applied to an image generated from a random latent code in $\mathcal{Z}$ space to obtain the target masked image. Additional results for Section 4.2 in the main document are shown in Fig. \ref{fig:fapp}. More results of randomly selected generated images after 500 iterations are shown in Fig. \ref{fig:fapp2}.\\

\begin{figure*}[p]
  \centering
   \includegraphics[width=1.0\linewidth]{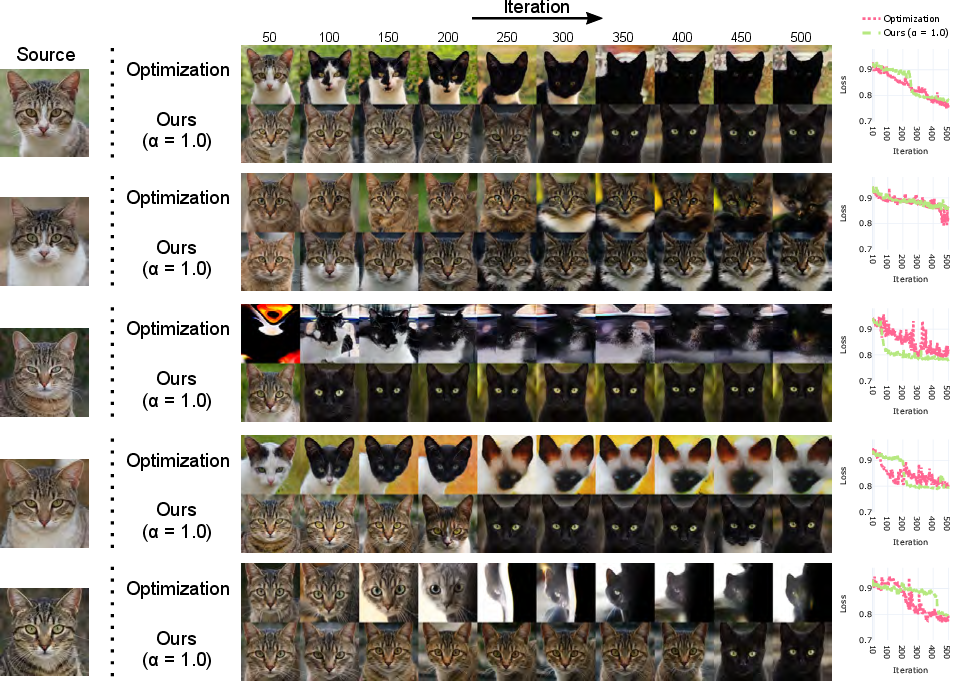}

   \caption{{\bf Text-guided Manipulation.} Latent codes for source images are manipulated toward the direction where the Contrastive Language-Image Pre-training (CLIP) embeddings for the generated images are similar to a CLIP embedding for a text prompt of ``a photo of a black cat.'' Compared with the baseline method, our method results in generated images that maintain photorealism.}
   \label{fig:capp}
\end{figure*}

\begin{figure*}[p]
  \centering
   \includegraphics[width=1.0\linewidth]{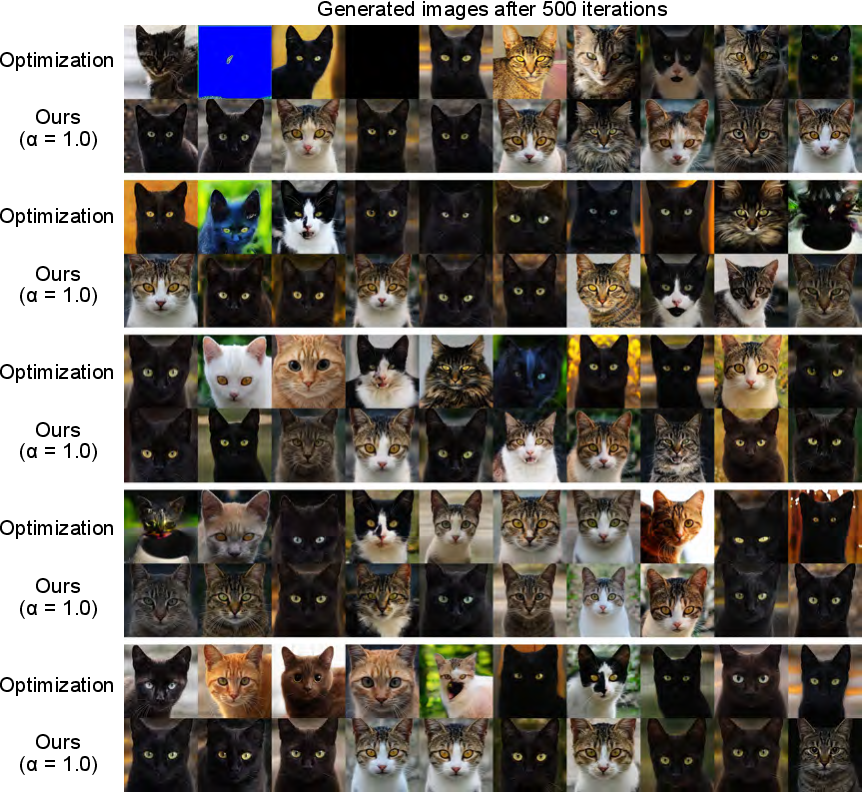}

   \caption{{\bf Text-guided Manipulation.} Randomly selected examples of generated images after 500 iterations.}
   \label{fig:capp2}
\end{figure*}

\noindent {\bf Text-guided Manipulation:} 
We use a Contrastive Language-Image Pre-training (CLIP) model pretrained on the LAION-400M dataset \cite{https://doi.org/10.48550/arxiv.2111.02114}. 
The codes and model weights for the CLIP model are available at \cite{oclip} (we used ``ViT-L-14, openai'') 
under the license \cite{oclc}. The LAION-400M dataset is available under the CC BY 4.0 license \cite{lalc}. Additional results for Section 4.2 in the main document are shown in Fig. \ref{fig:capp}. More results of randomly selected generated images after 500 iterations are shown in Fig. \ref{fig:capp2}.

\clearpage


{\small
\bibliographystyle{ieee_fullname}
\bibliography{egbib2}
}